\documentclass{article} % For LaTeX2e
\usepackage{colm2024_conference}

\usepackage{microtype}
\usepackage{hyperref}
\usepackage{url}
\usepackage{booktabs}
\usepackage{tcolorbox}
\tcbuselibrary{raster}
\tcbuselibrary{breakable}
\usepackage{graphicx}
\usepackage{xcolor}
\usepackage{tabularx}
\usepackage{multirow}
\usepackage{colortbl}
\usepackage{makecell}
\usepackage{xspace}
\usepackage{lineno}
\usepackage{emoji}
\usepackage{caption}
\usepackage{wrapfig}
\usepackage{longtable}
\usepackage{adjustbox}
\usepackage{listings}
\usepackage{lscape}

\definecolor{Box2Color}{rgb}{0.95,0.95,0.95}
\PassOptionsToPackage{table}{xcolor}
\title{Language Models as Compilers: Simulating Pseudocode Execution Improves Algorithmic Reasoning in Language Models}

\author{Hyungjoo Chae$^1$, Yeonghyeon Kim$^1$, Seungone Kim$^2$, Kai Tzu-iunn Ong$^1$, \\
\textbf{Beong-woo Kwak$^1$, Moohyeon Kim$^1$, Seonghwan Kim$^1$, Taeyoon Kwon$^1$,} \\
\textbf{Jiwan Chung$^1$, Youngjae Yu$^1$, Jinyoung Yeo$^1$} \\
\AND
$^1$Yonsei University \quad $^2$KAIST AI\\
\texttt{\{mapoout, jinyeo\}@yonsei.ac.kr} \\
}

\colmfinalcopy % Uncomment for camera-ready version, but NOT for submission.
\begin{document}

\maketitle
\newcommand{\todoc}[2]{{\textcolor{#1}{#2}}}
\newcommand{\todoblue}[1]{\todoc{blue}{#1}}

\newcommand{\todocc}[2]{{\textcolor{#1}{[[#2]]}}}
\newcommand{\todored}[1]{\todocc{red}{[[#1]]}}
\newcommand{\hist}[1]{\todored{hist: #1}}

\newcommand\codeprompt{$\mathcal{P}$\xspace}
\newcommand\nlprompt{$\mathcal{P}_{NL}$~}
\newcommand\donutemoji{\raisebox{-2pt}{\includegraphics[width=0.9em]{figure/doughnut.png}}}
\newcommand{\reasoner}{$\mathcal{R}$\xspace}
\newcommand{\instructor}{$\mathcal{I}$\xspace}
\newcommand{\machinecodeprompt}{$\mathcal{P}_{machine}$~}
\newcommand{\ours}{\textsc{Think-and-Execute}\xspace}

\newcommand{\todocccc}[2]{{\textcolor{#1}{[[#2]]}}}
\newcommand{\todogreen}[1]{\todocccc{green}{[[#1]]}}
\newcommand{\haejuu}[1]{\todogreen{haejuu: #1}}
\definecolor{mypink1}{rgb}{0.858, 0.188, 0.478}
\newcommand{\yeo}[1]{\textcolor{purple}{#1}}
\newcommand{\dragon}[1]{\textcolor{brown}{#1}}
\newcommand{\bwoo}[1]{\textcolor{olive}{#1}}
\newcommand{\kyle}[1]{\textcolor{blue}{#1}} 
\newcommand{\cheris}[1]{\textcolor{teal}{#1}}
\newcommand{\chatgpt}[1]{\textcolor{gray}{#1}}
\newcommand{\minus}[1]{\textcolor{red}{#1}}
\newcommand{\plus}[1]{\textcolor{ForestGreen}{#1}}
\newcommand{\person}{{$\mathbb{P}$}}
\newcommand{\thought}{$\mathbb{Z}_{\mathbb{P}}$}

\newcommand{\se}{{\it SE}}%
\newcommand{\eg}{{\it e.g.},~}%
\newcommand{\ie}{{\it i.e.},~}%
\newcommand{\etal}{{\it et al.}}%
\newcommand{\etc}{{\it etc}}%
\newcommand{\memory}{\mathcal{M}}%
\newcommand{\carecall}{$\text{CareCall}_{mem}$}
\newcommand{\acpm}{$\mathbbm{A}^2$-CPM}

\newcommand{\argmin}{\operatornamewithlimits{argmin}}
\newcommand{\argmax}{\operatornamewithlimits{argmax}}
\definecolor{yellow-green}{rgb}{0.3, 0.5, 0.0}

% \newcommand{\concat}{\DOTSB\concat@\slimits@}

% Caligraphy style
\newcommand{\mcal}[1]{{\cal{#1}}}
\newcommand{\calA}{\mbox{${\cal A}$}}
\newcommand{\calB}{\mbox{${\cal B}$}}
\newcommand{\calC}{\mbox{${\cal C}$}}
\newcommand{\calD}{\mbox{${\cal D}$}}
\newcommand{\calE}{\mbox{${\cal E}$}}
\newcommand{\calF}{\mbox{${\cal F}$}}
\newcommand{\calG}{\mbox{${\cal G}$}}
\newcommand{\calH}{\mbox{${\cal H}$}}
\newcommand{\calI}{\mbox{${\cal I}$}}
\newcommand{\calJ}{\mbox{${\cal J}$}}
\newcommand{\calK}{\mbox{${\cal K}$}}
\newcommand{\calL}{\mbox{${\cal L}$}}
\newcommand{\calM}{\mbox{${\cal M}$}}
\newcommand{\calN}{\mbox{${\cal N}$}}
\newcommand{\calO}{\mbox{${\cal O}$}}
\newcommand{\calP}{\mbox{${\cal P}$}}
\newcommand{\calQ}{\mbox{${\cal Q}$}}
\newcommand{\calR}{\mbox{${\cal R}$}}
\newcommand{\calS}{\mbox{${\cal S}$}}
\newcommand{\calT}{\mbox{${\cal T}$}}
\newcommand{\calU}{\mbox{${\cal U}$}}
\newcommand{\calV}{\mbox{${\cal V}$}}
\newcommand{\calW}{\mbox{${\cal W}$}}
\newcommand{\calX}{\mbox{${\cal X}$}}
\newcommand{\calY}{\mbox{${\cal Y}$}}
\newcommand{\calZ}{\mbox{${\cal Z}$}}

\definecolor{pythonblue}{rgb}{0.16,0.12,0.93}
\definecolor{cppgreen}{rgb}{0.16,0.42,0.16}
\definecolor{promptinsert}{HTML}{bfefff}
\definecolor{compcolor}{HTML}{90EE90}
\definecolor{codehlcolor}{HTML}{ffec8b}
\definecolor{codehlcolor2}{HTML}{ffbbff}
\definecolor{bgcolor}{rgb}{0.95,0.95,0.92}

% Styles for regular size code
\lstdefinestyle{python}{
    language=Python,
    basicstyle=\fontsize{8}{10}\ttfamily,
    keywordstyle=\color{blue},
    commentstyle=\color{gray},
    stringstyle=\color{black},
    showstringspaces=false,
    breaklines=true,
    breakindent=0pt,
    breakatwhitespace=false,
    escapeinside={(*@}{@*)}
}

\lstdefinestyle{cpp}{
    language=C++,
    basicstyle=\fontsize{8}{10}\ttfamily,
    keywordstyle=\color{blue},
    commentstyle=\color{gray},
    stringstyle=\color{green},
    showstringspaces=false,
    breaklines=true,
    breakindent=0pt,
    breakatwhitespace=false,
    escapeinside={(*@}{@*)}
}

% Small styles for examples in main text
\lstdefinestyle{plain}{
    basicstyle=\fontsize{8}{10}\ttfamily,
    keywordstyle=\color{blue},
    commentstyle=\color{gray},
    stringstyle=\color{green},
    showstringspaces=false,
    breaklines=true,
    breakatwhitespace=false,
    breakindent=0pt,
    escapeinside={(*@}{@*)}
}

\lstdefinestyle{python2}{
    language=Python,
    basicstyle=\fontsize{8}{10}\ttfamily,
    keywordstyle=\color{blue},
    commentstyle=\color{gray},
    stringstyle=\color{green},
    showstringspaces=false,
    breakatwhitespace=false,
    breaklines=true,
    breakindent=0pt,
    escapeinside={(*@}{@*)}
}

\lstdefinestyle{cpp2}{
    language=C++,
    basicstyle=\fontsize{8}{10}\ttfamily,
    keywordstyle=\color{blue},
    commentstyle=\color{gray},
    stringstyle=\color{green},
    showstringspaces=false,
    breaklines=true,
    breakindent=0pt,
    breakatwhitespace=false,
    escapeinside={(*@}{@*)}
}

\lstdefinestyle{sql}{
    language=SQL,
    basicstyle=\fontsize{8}{10}\ttfamily,
    keywordstyle=\color{blue},
    commentstyle=\color{green},
    stringstyle=\color{black},
    showstringspaces=false,
    breakatwhitespace=false,
    breaklines=true,
    breakindent=0pt,
    escapeinside={(*@}{@*)}
}

% Styles for prompts
% Prompt is for code, text is for regular text
\lstdefinestyle{prompt}{
    language=Python,
    basicstyle=\fontsize{8}{10}\ttfamily,
    keywordstyle=\color{blue},
    commentstyle=\color{gray},
    stringstyle=\color{cppgreen},
    showstringspaces=false,
    breaklines=true,
    backgroundcolor=\color{bgcolor},
    keepspaces=true, 
    breakindent=0pt,
    % linecolor=\color{lightgray},
    breakatwhitespace=false,
    showspaces=false,   
    escapeinside={(*@}{@*)}
}
\lstdefinestyle{text}{
    basicstyle=\fontsize{8}{10}\ttfamily,
    showstringspaces=false,
    breaklines=true,
    backgroundcolor=\color{bgcolor},
    breakatwhitespace=false,
    breakindent=0pt,
    keepspaces=true,
    showspaces=false,   
    escapeinside={(*@}{@*)}
}

\newcommand{\inserthl}[1]{\sethlcolor{promptinsert}\hl{#1}}
\newcommand{\comphl}[1]{\sethlcolor{compcolor}\hl{#1}}
\newcommand{\codehl}[1]{\sethlcolor{codehlcolor}\hl{#1}}
\newcommand{\codehlerr}[1]{\sethlcolor{codehlcolor2}\hl{#1}}

\definecolor{lightblue}{RGB}{224,236,247}
\definecolor{deepblue}{RGB}{9,46,107}

\begin{abstract}
Algorithmic reasoning refers to the ability to understand the complex patterns behind the problem and decompose them into a sequence of reasoning steps towards the solution. Such nature of algorithmic reasoning makes it a challenge for large language models (LLMs), even though they have demonstrated promising performance in other reasoning tasks.
Within this context, some recent studies use programming languages (\eg Python) to express the necessary logic for solving a given instance/question (\eg Program-of-Thought) as inspired by their strict and precise syntaxes.
However, it is non-trivial to write an executable code that expresses the correct logic on the fly within a single inference call. Also, the code generated specifically for an instance cannot be reused for others, even if they are from the same task and might require identical logic to solve.
This paper presents \ours, a novel framework that decomposes the reasoning process of language models into two steps. (1) In \textsc{Think}, we discover a task-level logic that is shared across all instances for solving a given task and then express the logic with \textit{pseudocode}; (2) In \textsc{Execute}, we further tailor the generated pseudocode to each instance and simulate the execution of the code.
With extensive experiments on seven algorithmic reasoning tasks, we demonstrate the effectiveness of \ours. Our approach better improves LMs' reasoning compared to several strong baselines performing instance-specific reasoning (\eg CoT and PoT), suggesting the helpfulness of discovering task-level logic. Also, we show that compared to natural language, pseudocode can better guide the reasoning of LMs, even though they are trained to follow natural language instructions.
\end{abstract}
\section{Introduction}
Reasoning in large language models (LLMs) typically entails analyzing the logical structure underlying a problem and realizing the logic into a sequence of reasoning steps to derive the final answer~\citep{zhou2022least,zhou2022teaching,Hao2023ReasoningWL}.
In particular, algorithmic reasoning has long been a formidable challenge for LLMs, as 
it requires to scrutinize a complicated reasoning pattern and to translate it into a long sequence of reasoning steps~\citep{suzgun2022challenging,valmeekam2022large,Pan2023LogicLMEL}.

To improve the reasoning capabilities of LLMs, prior works have primarily pursued two directions. The first direction includes enhancing the reasoning execution step by generating a rationale in natural language (\eg Chain-of-Thought~\citep{wei2022chain,kojima2022large}) or a piece of code (\eg Program-of-Thought~\citep{chen2023program}, Program-Aided LMs~\citep{gao2023pal}). However, such approaches perform step-by-step reasoning on-the-fly, without a dedicated phase for planning. This necessitates that the LLM analyze the logic and execute it within a single inference call, which constrains its expressiveness. Moreover, when encountering a similar problem, the LLM should solve it without being able to reuse the logic previously understood.

The second direction involves explicitly generating a plan described in natural language with LLMs. The plan describes the logic of the task and the LLM would subsequently concretize it into a sequence of reasoning steps (\eg Least-to-Most~\citep{zhou2022teaching}, Plan-and-Solve~\citep{wang-etal-2023-plan}). Yet, as prior works have mentioned, during our preliminary experiments, we find that natural language might not be the optimal medium to describe the logic of the problem~\citep{li2023chain}. In addition, prior works mostly rely on generating a plan by observing a single instance, which hinders analyzing the core reasoning pattern shared across similar instances within a single task~\citep{zhou2024self}. 

To address these issues, we introduce \ours, an algorithmic framework that discovers a logic that reflects the shared reasoning pattern behind a given task, and conducts reasoning by tailoring the logic into each instance. \ours consists of three distinctive steps; We first ask an LLM to \textsc{Think} about common reasoning patterns of a task by providing it with a few example questions. 
Then, the LLM translates the natural language description of the logic in a pseudocode format. 
The pseudocode format allows more flexibility in applying the logic to each instance compared to programming language such as Python.
Finally, in \textsc{Execute} step, the LLM simulates the execution of the task-level pseudocode to follow the logic in it and predicts the output result of the pseudocode.

Through extensive experiments on 7 algorithmic reasoning tasks from Big-Bench Hard~\citep{suzgun2022challenging}, we show the effectiveness of \ours over the challenging baselines.
The superior performance of \ours over PoT suggests that discovering the common logic for a given task and applying it to each instance would be more helpful than writing instance-specific code for every instance.
Noteworthily, simulating the execution of pseudocode is shown to improve LMs' reasoning more than planning with natural language (NL), even though they are trained to follow NL instructions.
Furthermore, we empirically show that the pseudocode prompt discovered by an LLM can be applied to small LMs (SLMs), such as CodeLlama-7B, to boost their reasoning ability. This indicates the efficiency of \ours over other code prompting methods that require the LLM to generate instance-specific code every time (\eg PoT). 
\begin{figure}
    \centering
    \includegraphics[width=0.95\linewidth]{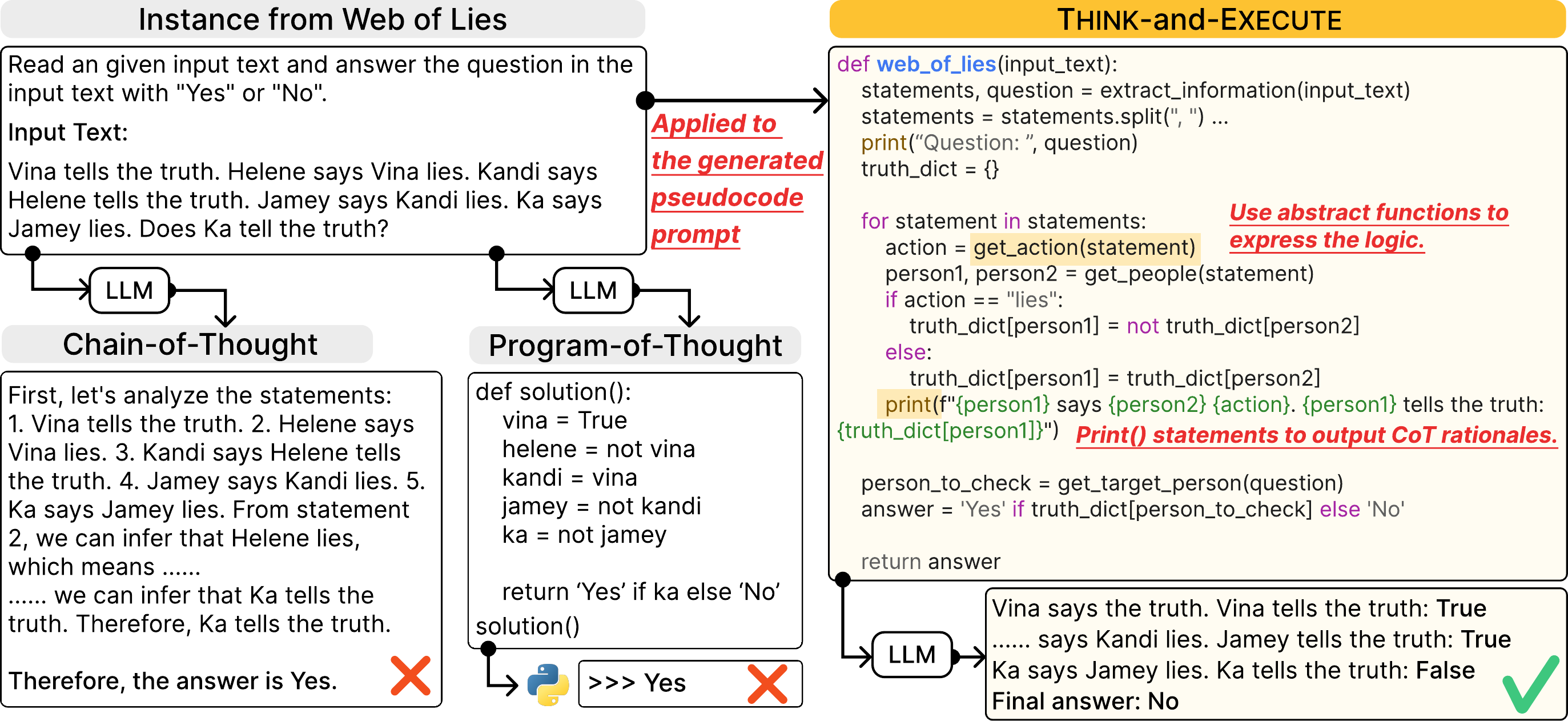}
    \caption{An illustration of \ours, compared with Zero-shot Chain-of-Thought~\citep{kojima2022large} and Program-of-Thoughts~\citep{chen2023program}.}
    \label{fig:comparison_with_prev_approaches}
\end{figure}

To summarize, our contributions are as follows:
\begin{itemize}
    \item We introduce \ours, a framework that performs reasoning with a pseudocode that contains the common logical structure of a given task.
    \item We show that \ours achieves notable improvements over strong baselines, including Chain-of-Thought and Program-of-Thought prompting, across various algorithmic tasks in Big-Bench Hard.
    \item We demonstrate that the pseudocode written by an LLM can be transferred to SLMs, showing the efficiency of our approach.
    
\end{itemize}

\section{\ours}
\begin{figure}
    \centering
    \includegraphics[width=0.9\linewidth]{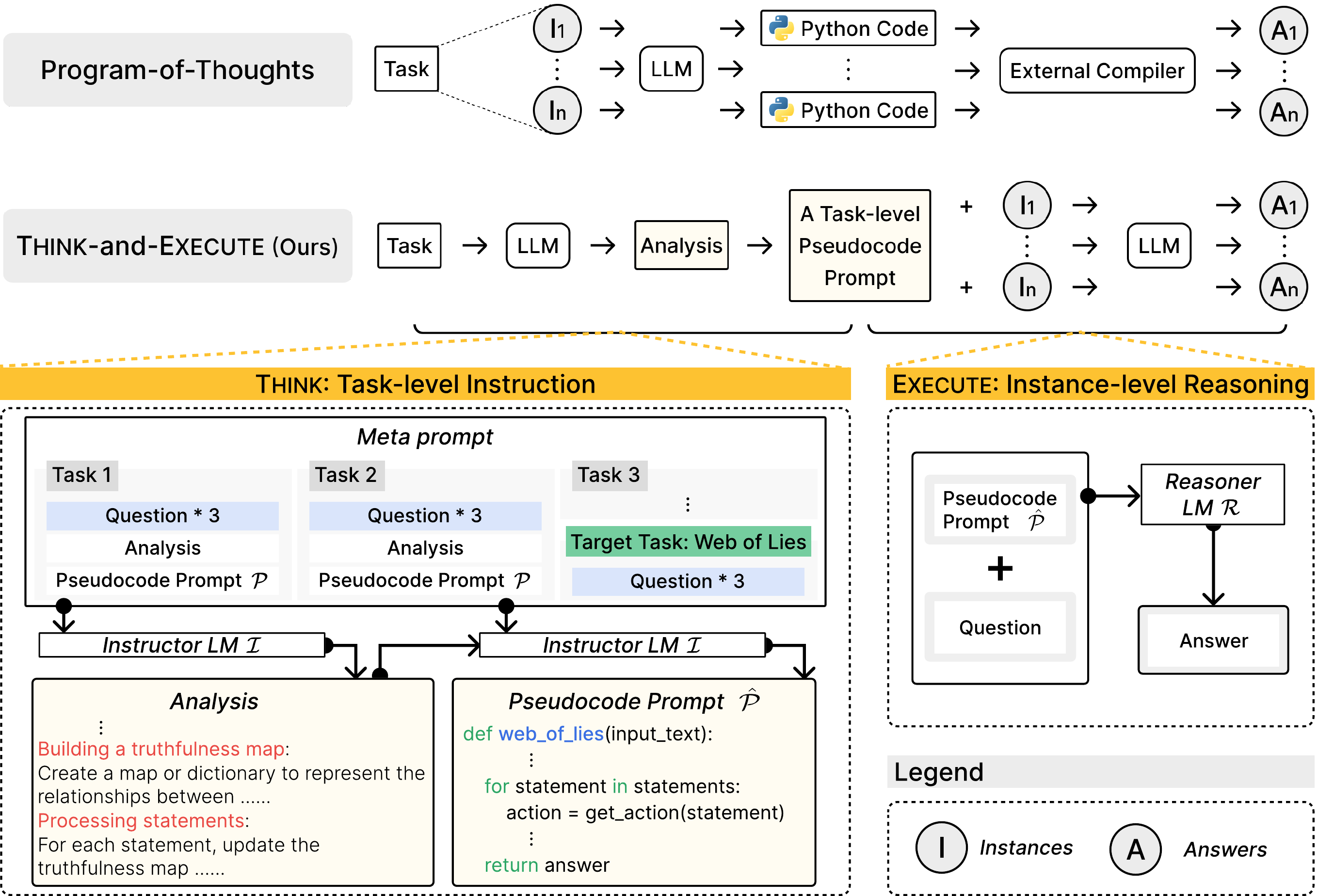}
    \caption{An overview of \ours. In \textsc{Think} (Top), an LLM analyzes the given task provided in the meta prompt and generates a pseudocode prompt that describes the necessary logic for solving the task. Then, in \textsc{Execute} (Bottom), the LLM conducts reasoning for each instance by simulating the execution of the pseudocode prompt.}
    \label{fig:overview}
\end{figure}

In this section, we introduce \ours and provide a detailed explanation of how LLMs perform reasoning with it. 
We incorporate an Instructor LM \instructor and a Reasoner LM \reasoner, for \textsc{Think} and \textsc{Execute}, respectively.
Figure~\ref{fig:overview} shows the overview of our framework.

\subsection{\textsc{Think}: Describing the Underlying Logic of a Task in a Pseudocode Format}
The goal for the Instructor LM \instructor in this phase is to discover the underlying logic for solving a given task $t$, and generate a prompt describing the logic, which will be further applied to all instances of the task (in \textsc{Execute}).
This prompt is constructed with \textbf{pseudocode} rather than natural language, which is used in prior work to guide the LM to perform step-by-step reasoning~\citep{kojima2022large, wang-etal-2023-plan}.

\paragraph{Step 1: Constructing a meta prompt.}
To prompt the Instructor LM \instructor to generate a task-level pseudocode for the given target task $t$, we provide \codeprompt of other tasks as demonstrations in a meta prompt.\footnote{We manually annotate \codeprompt for each task in $\mathcal{T}$ in advance. See Appendix~\ref{ssec:annotation} for examples.}
In practice, we construct the meta prompt with 3 randomly sampled tasks (3 example questions, analysis, and \codeprompt for each task) from $\mathcal{T}$ as demonstrations and the target task $t$ (3 example questions without the answers).\footnote{We use the questions of the examples instances in the few-shot prompt in Big-Bench Hard.}

\paragraph{Step 2: Analyzing the target task.}
Given the meta prompt, \instructor generates an analysis containing key reasoning logic that is required to solve the target task regardless of the instances (questions).
For example, in Figure~\ref{fig:overview} (Top), the generated analysis points out that \textit{building a truthfulness map} and updating it by \textit{processing statements} are needed to solve the task, \ie Web of Lies. This step guides \instructor to focus on the reasoning process shared among all the instances, which would be crucial in making a task-level prompt.
\paragraph{Step 3: Generating a pseudocode prompt based on the analysis.}
Next, based on the analysis, \instructor writes a prompt \codeprompt in the form of pseudocode, which breaks down the necessary reasoning steps for solving the target task.
We choose to use the pseudocode format over the form of natural language plan~\citep{kojima2022large, wang-etal-2023-plan} for two main reasons: (1) the efficiency of it in describing the logic behind a task (\eg avoid using repetitive instructions via \texttt{for} loop), and (2) the guidance of what and when to generate rationales via the argument in \texttt{print()} statement and the location within the execution of code.
For example, in Figure~\ref{fig:overview}, the \codeprompt contains the statement, {print(f"\{person1\} says \{person2\} \{action\}. \{person1\} tells the truth: \{truth\_dict[person1]\}")}, which instructs the Reasoner LM to generate a rationale that is helpful in keep tracking of the truth map containing the truthfulness of each person, during the execution of \codeprompt.
We provide more examples of meta prompt, analysis, and pseudocode prompt in Appendix~\ref{ssec:qual_analysis}.
\subsection{\textsc{Execute}: Simulating the Execution of Pseudocode Prompt for an Instance}
The reasoner LM \reasoner then conducts reasoning with the generated pseudocode prompt $\mathcal{P}$, tailoring the logic in \codeprompt for the given instance.
Following \citet{wei2022chain}, we aim to maximize the reasoning abilities of the LM by instructing them to explicitly generate intermediate reasoning steps, known as chain-of-thought (CoT) reasoning.
\reasoner is instructed to predict not only the final output result of the code, but also the intermediate execution outputs as rationales.
Specifically, \reasoner predicts a list of outputs $O = \{o_1, o_2, ..., o_k\}$ of the pseudocode by simulating the execution process of \codeprompt, where $o_i$ denotes the $i$-th system output from $\texttt{print()}$ statements, and $\{o_1\}_1^{k-1}$ are CoT rationales toward the final answer $o_k$.
We assume that tracking intermediate execution results would benefit \reasoner to keep track of the state of variables while they change over the execution of the code.
We enable \reasoner to mimic the behavior of a compiler with a system message ``\texttt{Generate the expected outputs (from all print() functions) of the code.}''.
% Formally, we regard the generation result of \reasoner as a list of system outputs $O = \{o_1, o_2, ..., o_k\}$, where $o_i$ denotes the $i$-th system output.
The final answer for a given question is outputted with ``\texttt{print("Final answer:\{answer\}")}'' command as the last system output $o_k$.

\section{Experimental Setup}

\subsection{Datasets} We curate seven algorithmic reasoning tasks from Big-Bench Hard~\citep{suzgun2022challenging}, including: dyck languages; geometric shapes; navigate; reasoning about colored objects; temporal sequence;tracking shuffled objectives; web of lies. These are specifically designed to measure the step-by-step reasoning capability of LLMs. Model performance on evaluated in \textbf{zero-shot} settings, where we do not provide demonstrations in the prompt. We provide detailed explanations in Appendix~\ref{ssec:bbh_explanation}.
% We also provide the results of other tasks of BBH that focus on other aspects of reasoning, such as usage of world knowledge and natural language understanding.

\subsection{Baselines}
We consider the following baselines: (1) \textbf{Direct prompting}: Directly predicting the answer without generating any rationales. (2) \textbf{Zero-shot CoT}~\citep{kojima2022large}: A setting where LLMs are evoked to generate the reasoning steps with ``\textit{Let's think step by step}'', before the answer. (3) \textbf{Zero-shot PoT}~\citep{chen2023program}: A setting where an LLM generates an instance-specific Python code that can be executed with a Python interpreter. Then, the execution result is used as the final answer. (4) \textbf{NL planning}: A variation of \ours, where the task-level instruction is generated in \textit{natural language}, instead of pseudocode.

% We compare our method in two different setting, hu
\subsection{Models}
For the Reasoner LM \reasoner, we adopt GPT-3.5-Turbo~\citep{openai2023chatgpt}, which shows strong performance in various reasoning benchmarks and code generation tasks~\citep{zellers-etal-2019-hellaswag,cobbe2021training, muennighoff2024octopack}, as well as the 7B and 13B versions of CodeLlama~\citep{roziere2023codellama}, which are trained on both code and natural language corpora and further fine-tuned to follow natural language instructions.
As for the Instructor LM \instructor, we choose GPT-3.5-Turbo.
% For the Reasoner LM \reasoner, we use 7B and 13B versions of models from the CodeLlama family~\citep{roziere2023codellama}, which are trained on both code and natural language corpora and further fine-tuned to follow natural language instructions. In addition, we use GPT-3.5-Turbo~\citep{openai2023chatgpt} which shows strong performance in various reasoning benchmarks~\citep{zellers-etal-2019-hellaswag,cobbe2021training} and code generation tasks~\citep{muennighoff2024octopack}.
% And we use GPT-3.5-Turbo for the Instructor LM \instructor.\footnote{We use \texttt{gpt-3.5-turbo-0125}.}

% \paragraph{Direct prompting with NL prompt.}
\section{Results}
\label{sec:results}
\newcommand{\specialcell}[2][c]{%
  \begin{tabular}[#1]{@{}c@{}}#2\end{tabular}}
\begin{table}
\centering
% \resizebox{1.\columnwidth}{!}{
\begin{tabular}{lccccccc||c}
\toprule
% \textbf{Method}            & \specialcell[]{boolean\\expression} & \specialcell{dyck\\languages} & \specialcell{geometric\\shapes} & \specialcell{logical\\deduction} & navigate & \specialcell{object\\counting} & \specialcell{colored\\objects} & \specialcell{temporal\\sequences} & \specialcell{tracking\\objectives} \\
\textbf{Reasoner/Method}            &  \specialcell{DL} & \specialcell{GS}  & Nav & \specialcell{CO}& \specialcell{TS} & \specialcell{SO} & \specialcell{WL} &  Avg\\
\midrule
\cellcolor{gray!17}\textit{CodeLlama-7B}        &    \cellcolor{gray!17}           &       \cellcolor{gray!17}        &     \cellcolor{gray!17}            &  \cellcolor{gray!17}                &  \cellcolor{gray!17}        &    \cellcolor{gray!17}             &   \cellcolor{gray!17}                                         &      \cellcolor{gray!17}                  \\
% - NL Prompt         &  &  &  &   &      &       &   & &          \\
Direct Prompting&0.0&9.0&39.0&\underline{24.4}&4.4&11.2&\underline{47.6}&19.4\\
Zero-shot CoT&0.0&\textbf{16.8}&26.0&10.8&\textbf{20.0}&10.4&44.8&18.4\\
NL Planning&0.0&10.0&\underline{52.0}&0.4&7.6&\underline{18.8}&\textbf{50.4}&\underline{19.9}\\
Zero-shot PoT&0.0&10.0&47.2&23.6&4.4&3.2&45.2&19.1\\
\ours&\textbf{2.0}&\underline{13.2}&\textbf{70.8}&\textbf{49.6}&\underline{19.2}&\textbf{22.0}&38.8&\textbf{30.8}\\
\midrule
\cellcolor{gray!17}\textit{CodeLlama-13B}        &    \cellcolor{gray!17}           &       \cellcolor{gray!17}        &     \cellcolor{gray!17}            &  \cellcolor{gray!17}                &  \cellcolor{gray!17}        &    \cellcolor{gray!17}             &   \cellcolor{gray!17}                                            &      \cellcolor{gray!17}                  \\
% - NL Prompt         &  &  &  &   &      &       &   & &          \\
Direct prompting&0.0&3.2&39.0&28.8&0.0&6.8&37.2&16.4\\
Zero-shot CoT&0.0&\textbf{24.8}&\underline{62.4}&28.0&\underline{21.6}&15.6&44.8&\underline{28.2}\\
NL Planning&\underline{1.2}&8.8&24.8&28.8&7.2&17.6&\textbf{53.6}&20.3\\
Zero-shot PoT&\underline{1.2}&16.4&45.6&\underline{38.8}&10.8&\textbf{35.6}&20.4&24.1\\
\ours&\textbf{8.0}&\underline{18.4}&\textbf{70.4}&\textbf{50.4}&\textbf{25.2}&\underline{32.4}&\underline{49.6}&\textbf{36.3}\\
\midrule
\cellcolor{gray!17}\textit{GPT-3.5-Turbo}              &       \cellcolor{gray!17}        &     \cellcolor{gray!17}            &  \cellcolor{gray!17}                &  \cellcolor{gray!17}        &    \cellcolor{gray!17}             &   \cellcolor{gray!17}                              &       \cellcolor{gray!17}               &      \cellcolor{gray!17}                  \\
% - NL Prompt         &  &  &  &   &      &       &   & &          \\
Direct prompting&1.0&33.0&57.0&\underline{52.4}&41.2&20.0&54.0&36.9\\
Zero-shot CoT&\underline{4.4}&\textbf{46.8}&73.2&70.4&\underline{44.4}&37.6&\underline{59.2}&\underline{48.0}\\
NL Planning&1.2&35.6&58.8&46.8&32.0&\underline{40.0}&50.4&37.8\\
Zero-shot PoT&0.4&21.2&\underline{77.2}&45.6&0.4&28.0&54.0&32.4\\
\ours&\textbf{6.0}&\underline{41.6}&\textbf{96.8}&\textbf{72.0}&\textbf{68.0}&\textbf{65.6}&\textbf{72.8}&\textbf{60.4}\\
\bottomrule
\end{tabular}
% }
\caption{Zero-shot performance of \ours and baselines on seven algorithmic reasoning tasks, including Dyck Languages (DL), Geometric Shapes (GS), Navigate (Nav), Reasoning about Colored Objects (CO), Temporal Sequences (TS), Tracking Shuffled Objectives (SO), and Web of Lies (WL) from Big-Bench Hard~\citep{suzgun2022challenging}.}
\label{tab:main_results}
\end{table}

% \subsection{Main Results}
\subsection{\ours Improves Algorithmic Reasoning}
We start by comparing our framework with direct prompting and zero-shot CoT~\citet{kojima2022large} in Table~\ref{tab:main_results}.
We find that zero-shot CoT performs better than direct prompting with average improvements of 11.1\% with GPT-3.5-Turbo, respectively, suggesting zero-shot CoT to be a strong baseline. 
Our \ours, however, further outperforms both of them significantly regardless of model sizes, which indicates that explicitly generating a plan is an effective way to improve the LLM's reasoning capabilities than simply encouraging LLMs to generate their intermediate reasoning steps.

\subsection{Task-level Pseudocode Prompts Benefits a Wider Range of Algorithmic Reasoning Tasks than Instance-specific Python Code}
In Table~\ref{tab:main_results}, PoT shows performance gains in some tasks over direct prompting (\eg Navigate; Tracking Shuffled Objects) with Python code generated specifically for each instance and the corresponding interpreter output as the answer.
However, such improvement is difficult to generalize to all tasks, \eg 0.4\% accuracy in both Dyck Language and Temporal Sequences, with GPT-3.5-Turbo.
% In Table~\ref{tab:main_results}, PoT shows performance gains in some tasks over direct prompting, such as Navigate and Tracking Shuffled Objects, by generating codes that are fully executable through real Python interpreters that are specially written to tackle a single instance.
% However, its benefit is difficult to extend to all tasks, such as Temporal Sequences, resulting in 0.4\% accuracy with GPT-3.5-Turbo.
By contrast, \ours outperforms PoT and direct prompting in all tasks with GPT-3.5-Turbo. This suggests that making the task-level strategy with pseudocode and applying it to each instance can benefit LLM's reasoning in a wider range of algorithmic reasoning tasks than generating instance-specific Python codes.
% On the other hand, \ours significantly outperforms PoT, showing a large amount of improvements over direct prompting for all tasks. This suggests that making a task-level reasoning strategy as a pseudocode prompt and applying it to each instance is more effective than writing instance-specific Python codes, even though they share some similar logic.

\subsection{The Logic Discovered by an LLM can be Transferred to SLMs}
We further explore if the pseudocode prompt written by an LLM (\ie GPT-3.5-Turbo as the instructor) can be applied to smaller LMs: the CodeLlama family in Table~\ref{tab:main_results}.
When applying the pseudocode prompts generated by GPT-3.5-Turbo, CodeLlama-7B and -13B significantly outperform direct prompting. Moreover, \ours with CodeLlama-13B shows comparable performance with GPT-3.5-Turbo with PoT and direct prompting.

\subsection{Pseudocode Better Describes the Logic for Solving a Task than Natural Language}

We also compare our approach with NL planning, a variant of ours that utilizes natural language to write the task-level instruction, instead of pseudocode. In practice, we provide human-written NL plans that contain a similar amount of information to \codeprompt in the meta prompt and use it to generate the task-level NL plan for the given task.
Surprisingly, although the LMs are fine-tuned to follow natural language instructions, we find that task-level pseudocode prompts can boost their performance more than NL plans (Table~\ref{tab:main_results}).

\subsection{Ablation Studies}
% \paragraph{Comparison with human-written pseudocode prompt.}
\begin{figure}[t]
    \centering
    \includegraphics[width=0.8\linewidth]{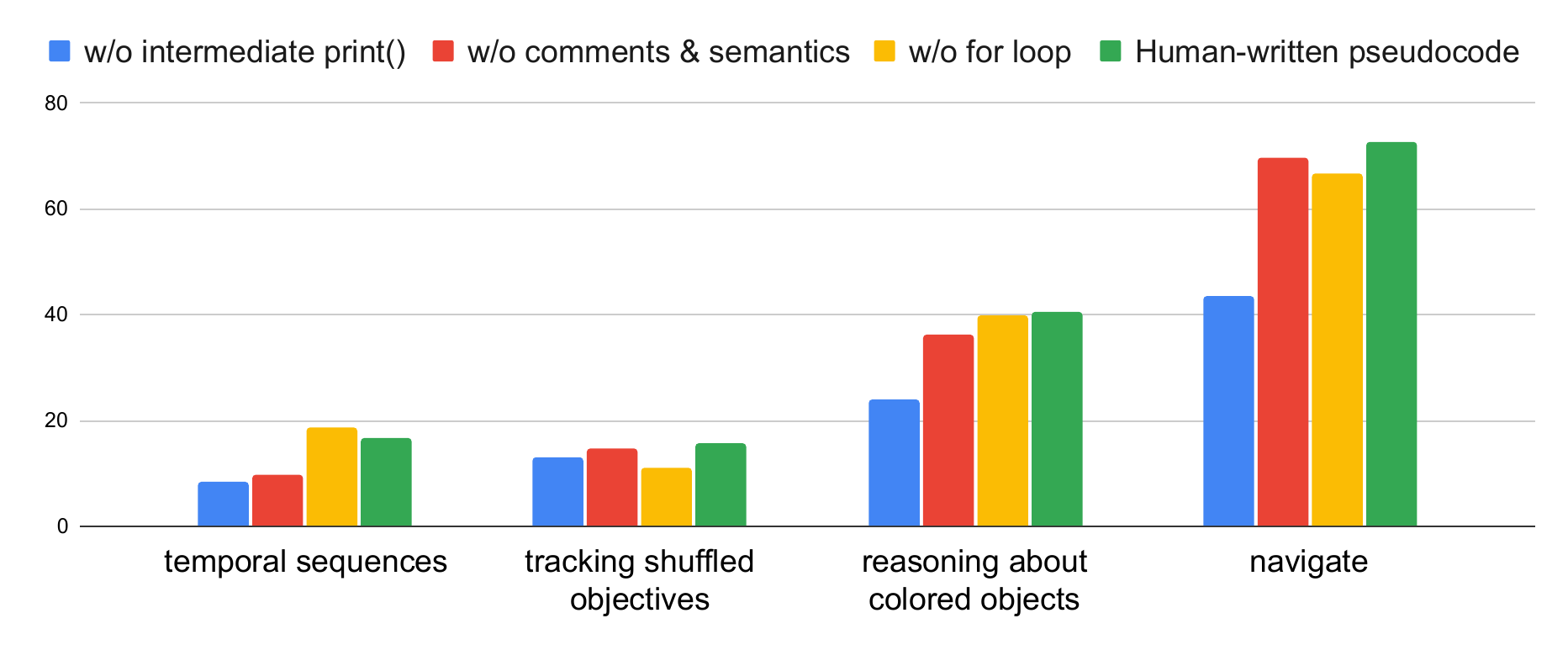}
    \caption{Ablation study of the components of pseudocode prompt using GPT-3.5-Turbo.}
    \label{fig:ablation}
\end{figure}

\paragraph{Components of the pseudocode prompt.}
We conduct an ablation study on each component of the pseudocode prompt.
For that, we prepare four types of pseudocode prompts: (1) \textbf{Human-written pseudocode}; (2) Human-written prompt \textbf{w/o comments and semantics} by removing the comments that explain the code and replacing variable names with meaningless alphabets, such as X, Y, and Z; (3) Human-written prompt \textbf{w/ for loop} and (4) \textbf{w/ intermediate print()} statements.
The results are in Figure~\ref{fig:ablation}. Model performance decreases significantly when applying prompts w/o comments and semantics, especially in Temporal Sequences. This implies that semantics play an important role in guiding the LLMs to apply the discovered logic and reasoning with it accordingly. 
Also, we find that printing out the intermediate execution steps with \texttt{print()} is crucial in reasoning, which is consistent with the finding from \citet{wei2022chain}.

\begin{table}[t]
\centering
\begin{tabular}{lc}
\toprule
\textbf{Method} & Avg \\
\midrule
w/o Analysis & 21.8 \\
\ours & \textbf{60.4}\\

\bottomrule
\end{tabular}
\caption{Ablation on Step2 of \textsc{Think} phase.}\label{tab:wo_analysis}
\end{table}

\paragraph{Generating the analysis before the pseudocode prompt.}
Table~\ref{tab:wo_analysis} shows a notable decrease in model performance when generating pseudocode prompts without conducting the analysis first. This suggests that explicitly generating analysis on the task can elicit a better pseudocode prompt that contains the necessary logic for solving the task.

\subsection{Comparison with other Baselines}

\begin{table}[t]
\centering
\begin{tabular}{cc}
\begin{tabular}{lc}
\toprule
\textbf{Method} & Avg  \\
\midrule
Chain-of-Code & 28.1 \\
Plan-and-Solve & 50.3 \\
\ours & \textbf{60.4}\\
\bottomrule
\end{tabular}
&
\begin{tabular}{lc}
\toprule
\textbf{Method} & Avg \\
\midrule
Self-Discover w/ GPT-4 & 77.9 \\
\ours w/ GPT-4& \textbf{81.7}\\
\bottomrule
\end{tabular}
\\
\end{tabular}
\caption{\textbf{Left}: Comparison of \ours, Chain-of-Code~\citep{li2023chain}, and Plan-and-Solve~\citep{wang-etal-2023-plan} using GPT-3.5-Turbo. \textbf{Right}: Comparison of \ours and Self-Discover~\citep{zhou2024self} using GPT-4. The results of Self-Discover are obtained from the original paper, as the code and prompts are not provided.}
\label{tab:other_baselines}
\end{table}

% \begin{table}[t]
% \centering
% \begin{tabular}{lc}
% \toprule
% \textbf{Method} & Avg \\
% \midrule
% Self-Discover w/ GPT-4 & 77.9 \\
% \ours w/ GPT-4& \textbf{82.8}\\

% \bottomrule
% \end{tabular}
% \caption{Comparison with Self-Discover~\citep{zhou2024self} using GPT-4.}\label{tab:self_discover}
% \end{table}

We further compare \ours with another three baselines: (1) Plan-and-Solve~\citep{wang-etal-2023-plan}, where an LLM sequentially generates a natural language plan for solving the given instance, step-by-step reasoning according to the plan, and the final answer; (2) Chain-of-Code~\citep{li2023chain}, where Python code is generated as a part of intermediate reasoning steps specifically for a given instance; (3) Self-Discover~\citep{zhou2024self}, a concurrent work that devises a task-level reasoning structure in a JSON format before inferencing the instance. 
First, as presented in Table~\ref{tab:other_baselines} (Left), we find \ours largely outperforms Plan-and-Solve and Chain-of-Code by 10.9 and 32.3 percentage points in terms of accuracy, respectively.
Second, while Self-Discover also incorporate task-level instruction, in Table~\ref{tab:other_baselines} (Right), our \ours with pseudocode prompts shows better performance when using GPT-4~\citep{achiam2023gpt}.\footnote{We use \texttt{gpt-4-0613} for GPT-4.} 
These findings indicate that generating (1) task-level instruction with (2) pseudocode can better represent the necessary logic for solving a task and benefit LLM's algorithmic ability. 
\section{Analysis}
\label{sec:analysis}
We conduct experiments to address the following research questions:
\begin{itemize}
    \item \textbf{RQ1}: Is task-level pseudocode more helpful than instance-specific pseudocode?
    \item \textbf{RQ2}: Does pre-training on code corpora improve reasoning?
    \item \textbf{RQ3}: How is the quality of the logic discovered by \ours compared to human-written logic? 
\end{itemize}

\subsection{Implementing the Underlying Logic is more Effective than Instance-specific Logic in Pseudocode (RQ1)}
We conduct an analysis to check if the improvement of \ours is contributed by our chosen format for the task-level instruction, \ie pseudocode. We compare \ours with a concurrent work, Chain-of-Code (CoC)~\citep{li2023chain}. 
In Table~\ref{tab:other_baselines}, \ours outperforms CoC, showing about 2x improvement in the average score.
The main difference between \ours and CoC is that we use pseudocodes which are generated to express logic shared among the tasks instances, while CoC incorporates pseudocode as part of the intermediate reasoning steps towards the solution of a given instance. Hence, the results indicate the advantages of applying pseudocode for the generation of task-level instruction over solely using them as a part of rationales.

% % \newcommand{\specialcell}[2][c]{%
%   % \begin{tabular}[#1]{@{}c@{}}#2\end{tabular}}
% \begin{table}[t]
% \centering
% % \resizebox{1.\columnwidth}{!}{
% \begin{tabular}{lccccccc||c}
% \toprule
% % \textbf{Method}            & \specialcell[]{boolean\\expression} & \specialcell{dyck\\languages} & \specialcell{geometric\\shapes} & \specialcell{logical\\deduction} & navigate & \specialcell{object\\counting} & \specialcell{colored\\objects} & \specialcell{temporal\\sequences} & \specialcell{tracking\\objectives} \\
% \textbf{Method}            &  \specialcell{DL} & \specialcell{GS}  & Nav & \specialcell{CO}& \specialcell{TS} & \specialcell{SO} & \specialcell{WL} &  Avg Acc\\
% \midrule
% % \cellcolor{gray!17}\textit{GPT-3.5-Turbo}              &       \cellcolor{gray!17}        &     \cellcolor{gray!17}            &  \cellcolor{gray!17}                &  \cellcolor{gray!17}        &    \cellcolor{gray!17}             &   \cellcolor{gray!17}                              &       \cellcolor{gray!17}               &      \cellcolor{gray!17}                  \\
% Chain-of-Code &2.8&17.6&57.2&26.0&16.8&29.6&46.4&28.1\\
% \ours&\textbf{6.0}&\textbf{41.6}&\textbf{96.8}&\textbf{72.0}&\textbf{68.0}&\textbf{65.6}&\textbf{72.8}&\textbf{60.4}\\
% \bottomrule
% \end{tabular}
% % }
% \caption{Comparison between \ours and Chain-of-Code~\citep{li2023chain} which generates pseudocodes specific to each instance.}
% \label{tab:instance_specific}
% \end{table}

\begin{figure}
    \centering
    \includegraphics[width=0.99\linewidth]{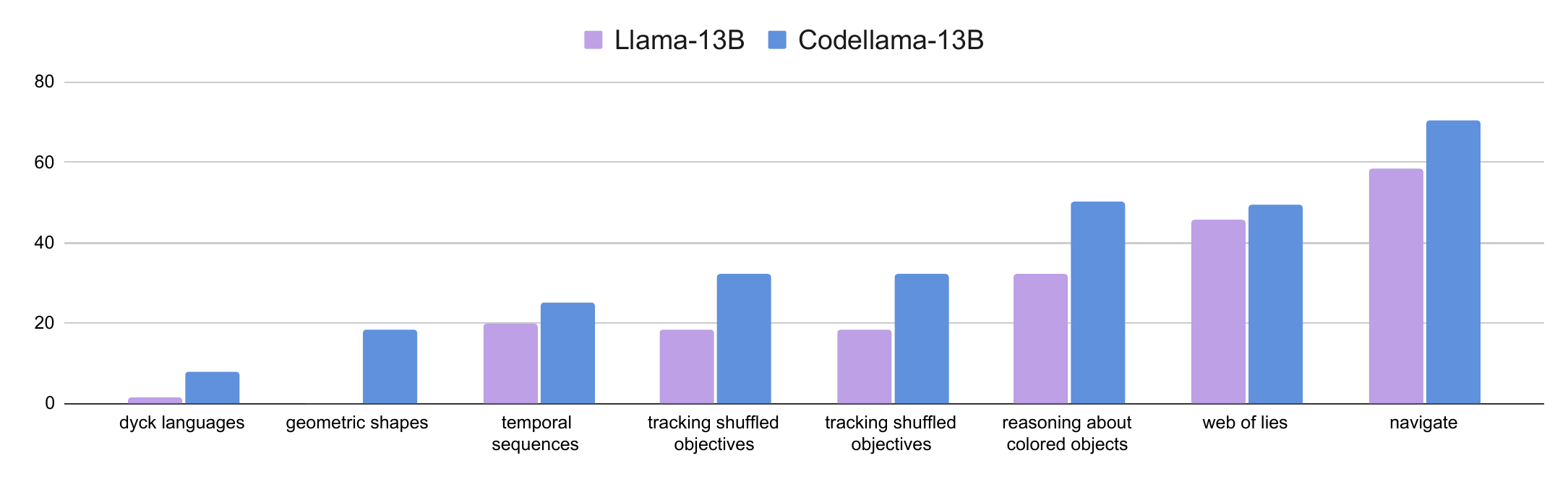}
    \caption{Analysis on the effect of code pre-training on the reasoning capability in applying \ours. Without pre-training on code corpora the accuracies drop notably.}
    \label{fig:enter-label}
    \label{fig:llama_vs_codellama}
\end{figure}
\subsection{\ours Requires Knowledge in Code (RQ2)}
To understand whether SLMs acquire the ability to understand the task-level logic written in pseudocode during pre-training on code corpora, we compare the performance of CodeLlama-13B with Llama-13B using \ours. In Figure~\ref{fig:llama_vs_codellama}, CodeLlama-13B shows better reasoning capabilities compared to Llama-13B in all tasks. These results suggest that the improvement from using \ours could depend on the knowledge of code, which is usually obtained by pre-training with code corpora. Writing code usually involves understanding the logic behind the given problem and expecting the execution results of a code, which resemble the same reasoning process of \ours.

\subsection{\ours can Generate a Logic Comparable to Human's (RQ3)}

\begin{table}[t]
\centering
% \resizebox{1.\columnwidth}{!}{
\begin{tabular}{lccccccc||c}
\toprule
% \textbf{Method}            & \specialcell[]{boolean\\expression} & \specialcell{dyck\\languages} & \specialcell{geometric\\shapes} & \specialcell{logical\\deduction} & navigate & \specialcell{object\\counting} & \specialcell{colored\\objects} & \specialcell{temporal\\sequences} & \specialcell{tracking\\objectives} \\
\textbf{Reasoner/Method}            &  \specialcell{DL} & \specialcell{GS}  & Nav & \specialcell{CO}& \specialcell{TS} & \specialcell{SO} & \specialcell{WL} &  Avg\\
\midrule
\cellcolor{gray!17}\textit{CodeLlama-7B}        &    \cellcolor{gray!17}           &       \cellcolor{gray!17}        &     \cellcolor{gray!17}            &  \cellcolor{gray!17}                &  \cellcolor{gray!17}        &    \cellcolor{gray!17}             &   \cellcolor{gray!17}                                         &      \cellcolor{gray!17}                  \\
Human-written \codeprompt&\textbf{2.4}&0.0&40.4&29.6&12.0&{18.0}&\textbf{52.8}&22.2\\
\ours&{2.0}&\textbf{13.2}&\textbf{70.8}&\textbf{49.6}&\textbf{19.2}&\textbf{22.0}&38.8&\textbf{30.8}\\
% - NL Prompt         &  &  &  &   &      &       &   & &          \\
\midrule
\cellcolor{gray!17}\textit{CodeLlama-13B}        &    \cellcolor{gray!17}           &       \cellcolor{gray!17}        &     \cellcolor{gray!17}            &  \cellcolor{gray!17}                &  \cellcolor{gray!17}        &    \cellcolor{gray!17}             &   \cellcolor{gray!17}                                            &      \cellcolor{gray!17}                  \\
Human-written \codeprompt&2.8&14.8&\textbf{72.8}&40.4&16.8&15.6&\textbf{49.6}&30.4\\
\ours&\textbf{8.0}&\textbf{18.4}&{70.4}&\textbf{50.4}&\textbf{25.2}&\textbf{32.4}&\textbf{49.6}&\textbf{36.3}\\
% - NL Prompt         &  &  &  &   &      &       &   & &          \\
\midrule
\cellcolor{gray!17}\textit{GPT-3.5-Turbo}              &       \cellcolor{gray!17}        &     \cellcolor{gray!17}            &  \cellcolor{gray!17}                &  \cellcolor{gray!17}        &    \cellcolor{gray!17}             &   \cellcolor{gray!17}                              &       \cellcolor{gray!17}               &      \cellcolor{gray!17}                  \\
% - NL Prompt         &  &  &  &   &      &       &   & &          \\
    Human-written \codeprompt&\textbf{12.4}&\textbf{50.0}&86.0&50.8&\textbf{84.0}&32.4&\textbf{74.4}&{55.7}\\

\ours&{6.0}&{41.6}&\textbf{96.8}&\textbf{72.0}&{68.0}&\textbf{65.6}&{72.8}&\textbf{60.4}\\
\bottomrule
\end{tabular}
% }
\caption{Comparison between \ours and Human-written \codeprompt.}
\label{tab:human_vs_ours}
\end{table}
To gauge LLMs' capabilities in discerning the underlying logic of a task, we compare \ours (using GPT-3.5-Turbo as the Instructor) with human-written pseudocode prompts. The results are shown in Table~\ref{tab:human_vs_ours}. Using the GPT-3.5-Turbo the Reasoner, \ours scores 60.4\% in terms of accuracy, which is superior to the human-written \codeprompt (with an accuracy of 55.7\%). Especially, in the tasks of Navigate and Tracking Shuffled Objectives, pseudocode prompts generated by \ours elicit better performance. This also holds true when adopting CodeLlama-7B and -13B as the Reasoner, further suggesting the effectiveness of our \textsc{Think} step over human writers.

%Please add the following packages if necessary:
%\usepackage{booktabs, multirow} % for borders and merged ranges
%\usepackage{soul}% for underlines
%\usepackage[table]{xcolor} % for cell colors
%\usepackage{changepage,threeparttable} % for wide tables
%If the table is too wide, replace \begin{table}[!htp]...\end{table} with
%\begin{adjustwidth}{-2.5 cm}{-2.5 cm}\centering\begin{threeparttable}[!htb]...\end{threeparttable}\end{adjustwidth}

\begin{table}[t]
\centering
\begin{tabular}{lcccc}
\toprule
\multirow{2}{*}{\textbf{Reasoner}} & \multicolumn{3}{c}{\textbf{Instructor}} \\
\cmidrule(r){2-4}
 & CodeLlama-13B & CodeLlama-34B &GPT-3.5-Turbo \\
\midrule
CodeLlama-13B & 30.9 & 33.0 & 36.4 \\
CodeLlama-34B & 32.5 & 34.2 & 39.1\\
GPT-3.5-Turbo & 33.9 & 35.9 & 60.4 \\
\bottomrule
\end{tabular}
\caption{Analysis of the effect of the capability of Reasoner and Instructor on the performance. We report the average performance on the 7 tasks.}\label{tab:reasoner_comparison}
\end{table}

\subsection{Impact of LLMs' Capability on \ours}
In examining the impact of LLMs' capabilities within our framework, we investigate the influence of both the Reasoner and Instructor components on performance, as depicted in Table~\ref{tab:reasoner_comparison}. Notably, higher accuracy scores are observed when utilizing GPT-3.5-Turbo as Reasoners compared to CodeLlama-13B and CodeLlama-34B. Additionally, the effectiveness of the Instructor also plays a crucial role, with GPT-3.5-Turbo exhibiting the highest accuracy scores across all configurations. These results underscore the significance of both the Reasoner and Instructor components in enhancing the performance of \ours.
\section{Related Work}
% kai: i can write this part if you want.

\paragraph{Chain-of-Thought prompting.}
Chain-of-thought (CoT) prompting evokes LMs to generate intermediate reasoning steps that guide and explain the solution~\citep{wei2022chain, wang2022self, wu-etal-2023-chain}. One common paradigm of this is zero-shot CoT prompting~\citep{kojima2022large}. Without specifically designed question-explanation-answer triplets as demonstrations, zero-shot CoT prompting elicits a plausible reasoning path towards the final answer with simple instruction, such as \textit{"Let's think step-by-step"}, eliciting better model performance in tasks that require multi-step reasoning.

In the context of improving zero-shot CoT, \citet{wang-etal-2023-plan} propose to first generate a plan breaking down the target task into smaller subtasks, and then solve each subtask according to the plan. Similar to our approach, a concurrent work~\citep{zhou2024self} devises a task-level reasoning structure that can be applied to each instance (question) of the target task. The most significant distinction between these prior studies and ours is that our \ours adopts \textbf{pseudocode} (as opposed to natural language) to express the necessary logic for solving the task. We demonstrate that our task-level pseudocode prompt empowers LMs with better ability of zero-shot reasoning than natural language plans under various settings in Section~\ref{sec:analysis}.

\paragraph{Incorporation of code in reasoning.}
With unambiguous syntaxe and strict structure, programming languages such as Python have been applied to LLM-based systems to improve system performance in solving tasks. For instance, \citet{gao2023pal} and \citep{chen2023program} use LLMs to generate Python code for given mathematical questions, and run the generated code on external compilers to obtain/calculate the answers. Concurrently with our work, \citet{li2023chain} present chain-of-code (CoC), where pseudocode is also incorporated along with the Python code for solving a given question (instance). While this approach generates instance-specific code as intermediate reasoning steps for each individual instance, our \ours, by contrast, focus on the task-level pseudocode prompt that can be applied to all instances. We compare CoC and \ours in Section~\ref{sec:results}.
\section{Limitations and Discussion}
A possible limitation of our approach is that we focus on algorithmic reasoning, as we believe it is the best setting to assess LLMs' capabilities in understanding a complex logic and carrying out a sequence of reasoning step, following the logic.
However, we believe that \ours can be applied to other domains of reasoning that require following a long sequence of reasoning steps, such as multi-hop reasoning~\citep{Ji2020LanguageGW} and symbolic reasoning~\citep{madaan2022text}.

\section{Conclusion}
In this paper, we present \ours, an algorithmic reasoning framework that generates a logic for solving the given task into a pseudocode and performs reasoning by simulating the execution of the pseudocode with language models.
Through extensive experiments, we show the effectiveness of \ours, over the strong baselines.
These results underscore not only the usefulness of pseudocode in eliciting language models' reasoning capabilities but also the efficiency of our framework in discovering the high-quality logic behind a given task.



% \bibliography{anthology,colm2024_conference}
\bibliography{_colm2024_conference}
\bibliographystyle{colm2024_conference}
\clearpage
\appendix

\section{Experimental Details}

\subsection{Models}
% Please specify the url link to download the checkpoint if applicable. Otherwise, metion the api service.
We use several LLMs, including GPT-3.5-Turbo \citep{openai2023chatgpt} and GPT-4 \citep{achiam2023gpt}, which are available via OpenAI API\footnote{\url{https://openai.com/blog/openai-api}}, and open-source LLM, CodeLlama \citep{roziere2023codellama} as the Instructor LM $\mathcal{I}$ and the Reasoner LM $\mathcal{R}$.
\begin{itemize}
    \item \textbf{GPT-3.5-Turbo}: \texttt{gpt-3.5-turbo-0125}
    \item \textbf{GPT-4}: \texttt{gpt-4-0613}
    \item \textbf{CodeLlama}: CodeLlama encompasses variations of LLaMA2 fine-tuned for code domains using code corpus. 
    This comprehensive collection features models of various sizes (7B, 13B, 34B, and 70B) and diverse types, including the foundation model, Python-focused model, and instruction-following model. 
    In our study, we employ the CodeLlama-Instruct model (7B\footnote{\url{https://huggingface.co/codellama/CodeLlama-7b-Instruct-hf}}, 13B\footnote{\url{https://huggingface.co/codellama/CodeLlama-13b-Instruct-hf}}).
\end{itemize}

\subsection{Inference}
% vLLM, GPT specification. Amount of time required to inference all the instances (for GPT-3.5-Turbo and CodeLlama, respectively).
We use vLLM to improve inference throughput.\footnote{\url{https://github.com/vllm-project/vllm}}
During our experiments, we adopt temperature sampling with $T=0.0$ (\ie greedy decoding) to efficiently generate outputs.
For a task comprising 250 instances, GPT-3.5-Turbo achieves an inference time of 30 seconds.
Additionally, utilizing 2 A100 GPUs, CodeLlama achieves inference times of approximately 2 and 5 minutes for 7B and 13B models, respectively.
% ChatGPT -> 30초 내외, codelllama는 2xA100으로 7B, 13B 각각 2min, 5min 정도

\subsection{Evaluation}
% Answer triggers, use of accuracy following BBH.
To extract answers for evaluation, LLMs generate the final answer triggered by the phrase "Final answer: ".
Following~\citet{suzgun2022challenging}, we provide all multiple-choice options to LLMs as input, then measure accuracy using exact match (EM), which compares the generated output with the ground-truth label. To ensure fair comparison between PoT and other baselines, we also admit the prediction that includes the text of correct choice, \eg blue, but without a choice tag, \eg "(A)".

\subsection{Datasets}
We take 7 algorithmic benchmarks from Big-Bench Hard~\citep{suzgun2022challenging} dataset. All datasets contain 250 examples respectively. We provide the descriptions of each dataset regarding the goals and contexts.
\begin{itemize}
    \item \textbf{Dyck Languages (DL)}: Complete a partially given Dyck-4 sequence by predicting the necessary sequence of closing brackets that are missing at the end.
    \item \textbf{Geometric Shapes (GS)}: Determine the geometric figure formed by following all the instructions in a specified SVG path element containing several commands.
    \item \textbf{Navigate (Nav)}: Evaluate whether a set of directional commands will return a navigator to the starting point.
    \item \textbf{Reasoning about Colored Objects (CO)}: Given a scenario, deduce the color of a specific object placed on a surface, using the provided context for guidance. 
    %  please 
    \item \textbf{Temporal Sequences (TS)}: Examine a chronology of a person’s daily activities to find when they could fit an additional activity into their schedule.
    \item \textbf{Tracking Shuffled Objectives (SO)}: Ascertain the final positions of several objects after they have been moved from their original locations through a sequence of exchanges. We use the version of the task with 5 objectives.
    \item \textbf{Web of Lies (WL)}: Assess the veracity of a Boolean function presented within a narrative problem to establish its truthfulness.
\end{itemize}

\label{ssec:bbh_explanation}

\section{Details of \ours}
\subsection{Human-annotation on the Tasks in the Task Pool}
Please see Appendix~\ref{ssec:human-written} for human-written pseudocode prompts.
\label{ssec:annotation}
\subsection{Components of a Pseudocode Prompt}
We highlight some components of code prompt that would be helpful in describing the underlying reasoning logic. 
\begin{itemize}
\item \textbf{Conditional branch}:  
To allow the reasoning model to take different reasoning paths based on the condition, we use \texttt{if} and \texttt{else} statement to describe the logic.

\item\textbf{Loop}: We can efficiently present repetitive instructions that iterate over a list of items by using loops, such as \texttt{for} and \texttt{while} loop.

\item \textbf{Abstraction}: 
In programming, we can encapsulate a complex logic into a single function.
% Abstraction in programming refers to the process of encapsulating complex details, making it simpler to understand and use. 
Focusing on this, we adopt modular design in constructing pseudocode prompts by encapsulating complex and repetitive process into an abstract function.

\item \textbf{Variables}: Variables are essential in programming languages as they store data values to execute instructions. Similarly, in reasoning, keeping track of variables is crucial for maintaining state, passing data, and for general data manipulation tasks. 

\item \textbf{Comments and docstrings}: As human programmers can rely on the assistance of comments to better understand codes, we provide more detailed explanations on the intent of code via comments. Also, comments and docstrings can compensate the limitation when some semantics cannot be directly expressed with programming language.
\end{itemize}

\subsection{Comparison to Related Work}

Table~\ref{tab:related_work_short_summary} summarizes some related approaches to ours.
\begin{table}[!ht]  
    \centering
    \resizebox{1.0\columnwidth}{!}{%
    \begin{tabular}{lccc}
    \toprule
    \textbf{Method} & \textbf{Granularity of plan/logic} & \textbf{Use of pseudocode} & \textbf{Transferability to SLMs}  \\ \midrule
    Plan-and-Solve~\citep{wang-etal-2023-plan}
    & Instance-level & \emoji[emoji]{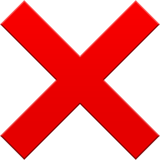}  & \emoji[emoji]{x} \\
    % \\ \midrule
    Self-Discover~\citep{zhou2024self} 
    & Task-level & \emoji[emoji]{x}   & \emoji[emoji]{x} \\
    Chain-of-Code~\citep{li2023chain} 
    & Intance-level & \emoji[emoji]{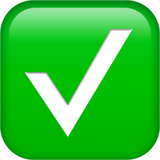}   & \emoji[emoji]{x} \\
    % \\ \midrule
    \textbf{\ours} (this work) 
    & Task-level & \emoji[emoji]{check} & \emoji[emoji]{check} 
    \\ \bottomrule
    \end{tabular}%
    }
    \caption{A comparison of \ours to closely related prior approaches.}
    \label{tab:related_work_short_summary}
\end{table}

% We provide the generated analyes for each task in Figure~\ref{fig:analysis:dyck_languages}, ~\ref{fig:analysis:geometric_shapes}, ~\ref{fig:analysis:navigate}, ~\ref{fig:analysis:temporal_sequences}, ~\ref{fig:analysis:tracking}, \ref{fig:analysis:colored_objects}, and ~\ref{fig:anaylsis:web_of_lies},

% We provide the generated pseudocode prompts for each task in Figures~\ref{fig:generated_prompt:dyck_languages}, ~\ref{fig:generated_prompt:geometric_shaped}, ~\ref{fig:generated_prompt:naviate}, ~\ref{fig:generated_prompt:temporal_sequences}, ~\ref{fig:generated_prompt:colored_objects}, ~\ref{fig:generated_prompt:tracking}, and~\ref{fig:generated_prompt:web_of_lies}.
\section{Prompts Used in Our Experiments}
\subsection{Meta Prompt for generating an analysis (\textsc{Think}: Step 2).}
\begin{tcolorbox}[breakable, toprule at break=0pt, bottomrule at break=0pt,colback=white]
\begin{lstlisting}[style=text, columns=fullflexible]
Generate an explanation, analyzation, and plan to generate code prompt for the last task considering the example task instances. Your plan should show enough intermediate reasoning steps towards the answer. Construct the plan as much as you can and describe the logic specifically. When constructing the plan for the code prompt, actively use 'if else statement' to take different reasoning paths based on the condition, 'loop' to efficiently process the repititive instructions, 'dictionary' to keep track of connections between important variables.

[Example 1]
Example task instances:
{example_instances_of_task1}

Output format:
{output_format_of_task1}

Explanation:
{analysis_of_task1}

...

[Example 4]
Example task instances:
{example_instances_of_target_task}

Output format:
{output_format_of_target_task}

Explanation:
\end{lstlisting}
\end{tcolorbox}
%     \caption{Prompt for generating an analysis (\textsc{Think}: Step 2).}
%     \label{fig:prompt:analysis}
% \end{figure}

% \subsection{Prompt for generating code prompt}
\subsection{Meta Prompt for pseudocode prompt genration (\textsc{Think}: Step 3).}
\begin{tcolorbox}[breakable, toprule at break=0pt, bottomrule at break=0pt,colback=white]
\begin{lstlisting}[style=text, columns=fullflexible]
Generate the code prompt for the last task using the similar style of the example codes. Add enough print() functions following the provided steps  in the provided explanation to output intermediate reasoning steps towards the answer and keep track of important variables. Implement the code prompt as much as you can and describe the logic in code following the provided explanation but do not make a code that is biased toward a single task example instance. For example, do not use hard-coded variables that are obtained from task instances (e.g., using specific name of person in the question). The code prompt must be able to be applied to various instances of same task. When returning the final answer, carefully consider the output format. Especially, for the multiple choice questions, the final answer should be one of the given options. The main function name should be '{function_name}'. Along with the main function, you may want to define some helper functions that might be helpful for implementing the '{function_name}'. But you don't have to explicitly implement the helper functions, but just define them with function name and a single-line explanation in comment. When constructing the main function, ...

[Example 1]
Task description:
{description_of_task1}

Example task instances and the code usage:
{example_task_instances_and_code_usages_of_target_task}

Format of the Final answer:
{output_format_of_task1}

Explanation:
{analysis_of_task1}

Code prompt:
{code_prompt_of_task1}

...

[Example 4]
Task description:
{description_of_target_task}

Example task instances and the code usage:
{example_task_instances_and_code_usages_of_target_task}

Format of the Final answer:
{output_format_of_target_task}

Explanation:
{analysis_of_target_task}

Code prompt:
\end{lstlisting}
\end{tcolorbox}

% \begin{figure}
%     \begin{small}
%         \captionsetup{justification=centering, labelfont=bf, font=small}
%         \begin{minted}[fontsize=\footnotesize, frame=lines, framesep=2mm, baselinestretch=1.2, breaklines, breaksymbolleft={}, breaksymbolright={},bgcolor=Box2Color]{text}
% Generate a plan for the last task considering the example task instances. Your plan should show enough intermediate reasoning steps towards the answer. Construct the plan as much as you can and describe the logic specifically.

% [Example 1]
% Task description:
% {description_of_task1}

% [Example 1]
% Example task instances:
% {example_instances_of_task1}

% Output format:
% {output_format_of_task1}

% Plan:
% {analysis_of_task1}

% ...

% [Example 4]
% Example task instances: {example_instances_of_target_task}

% Output format:
% {output_format_of_target_task}

% Plan:
%         \end{minted}
%     \end{small}
%     \caption{Prompt for NL Planning (\textsc{Think}: Step 3).}
%     \label{fig:prompt:nl_planning}
% \end{figure}

\subsection{Prompt for NL Planning}
\begin{tcolorbox}[breakable, toprule at break=0pt, bottomrule at break=0pt,colback=white]
\begin{lstlisting}[style=text, columns=fullflexible]
Generate a plan for the last task considering the example task instances. Your plan should show enough intermediate reasoning steps towards the answer. Construct the plan as much as you can and describe the logic specifically.

[Example 1]
Task description:
{description_of_task1}

[Example 1]
Example task instances:
{example_instances_of_task1}

Output format:
{output_format_of_task1}

Plan:
{analysis_of_task1}

...

[Example 4]
Example task instances: {example_instances_of_target_task}

Output format:
{output_format_of_target_task}

Plan:
\end{lstlisting}
\end{tcolorbox}

% \subsection{Code prompt scoring template}

% \begin{figure}
%     \begin{small}
%         \captionsetup{justification=centering, labelfont=bf, font=small}
%         \begin{minted}[fontsize=\footnotesize, frame=lines, framesep=2mm, baselinestretch=1.2, breaklines, breaksymbolleft={}, breaksymbolright={},bgcolor=Box2Color]{text}
% {prompt}
% input_text = "{input_text}"
% final_answer = {function_name}(input_text)
% print("Final answer:"+ final_answer)
% Generate the expected execution output (output from all print() functions) of the code. You don't have to actually run the code and do not care about 'not implemented error'.
% Expected output:
%         \end{minted}
%     \end{small}
%     \caption{Prompt for \textsc{Execute} phase.}
%     \label{fig:prompt:execute}
% \end{figure}

\subsection{Prompt for \textsc{Execute} phase}
\begin{tcolorbox}[breakable, toprule at break=0pt, bottomrule at break=0pt,colback=white]
\begin{lstlisting}[style=text, columns=fullflexible]
{prompt}
input_text = "{input_text}"
final_answer = {function_name}(input_text)
print("Final answer:"+ final_answer)
Generate the expected execution output (output from all print() functions) of the code. You don't have to actually run the code and do not care about 'not implemented error'.
\end{lstlisting}
\end{tcolorbox}

\subsection{Prompt for evaluating \textbf{Direct Prompting}}
\begin{tcolorbox}[breakable, toprule at break=0pt, bottomrule at break=0pt,colback=white]
\begin{lstlisting}[style=text, columns=fullflexible]
{prompt}
text for the task: {input_text}
Final answer should be at the end of your answer and its format should be like "Final answer: your_answer".
Generate output following the task description above.
Output:
\end{lstlisting}
\end{tcolorbox}

% \subsection{Zero-shot CoT scoring template}

\subsection{Prompt for evaluating \textbf{Zero-shot CoT}}
\begin{tcolorbox}[breakable, toprule at break=0pt, bottomrule at break=0pt,colback=white]
\begin{lstlisting}[style=text, columns=fullflexible]
{prompt}
text for the task: {input_text}
Final answer should be at the end of your answer and its format should be like "Final answer: your_answer".
Generate output following the task description above.
Output:
Let's think step by step.
\end{lstlisting}
\end{tcolorbox}

% \begin{figure}
%     \begin{small}
%         \captionsetup{justification=centering, labelfont=bf, font=small}
%         \begin{minted}[fontsize=\footnotesize, frame=lines, framesep=2mm, baselinestretch=1.2, breaklines, breaksymbolleft={}, breaksymbolright={},bgcolor=Box2Color]{text}
% You will write python program to solve the below problem. You will only write code blocks. Your python promgram must be executable and returns the right answer for the problem.

% Q: {question}

% # solution using Python:

% def solution():
%     """{question}"""
%         \end{minted}
%     \end{small}
%     \caption{Prompt for evaluating \textbf{Zero-shot PoT}}
%     \label{fig:prompt:pot}
% \end{figure}

% \subsection{Generated Analysis of Geometric Shapes}
\subsection{Prompt for evaluating \textbf{Zero-shot PoT}}
\begin{tcolorbox}[breakable, toprule at break=0pt, bottomrule at break=0pt,colback=white]
\begin{lstlisting}[style=text, columns=fullflexible]
You will write python program to solve the below problem. You will only write code blocks. Your python promgram must be executable and returns the right answer for the problem.

Q: {question}

# solution using Python:

def solution():
    """{question}"""
\end{lstlisting}
\end{tcolorbox}

% \subsection{Plan-and-Solve scoring template}

% \begin{figure}
%     \begin{small}
%         \captionsetup{justification=centering, labelfont=bf, font=small}
%         \begin{minted}[fontsize=\footnotesize, frame=lines, framesep=2mm, baselinestretch=1.2, breaklines, breaksymbolleft={}, breaksymbolright={},bgcolor=Box2Color]{text}
% {prompt}
% text for the task: {input_text}
% Final answer should be at the end of your answer and its format should be like "Final answer: your_answer".
% Generate output following the task description above.
% Output:
% Let’s first understand the problem and devise a plan to solve the problem. Then, let’s carry out the plan and solve the problem step by step.
%         \end{minted}
%     \end{small}
%     \caption{Prompt for evaluating \textbf{Plan-and-Solve}}
%     \label{fig:prompt:plan_and_solve}
% \end{figure}

\subsection{Prompt for evaluating \textbf{Plan-and-Solve}}
\begin{tcolorbox}[breakable, toprule at break=0pt, bottomrule at break=0pt,colback=white]
\begin{lstlisting}[style=text, columns=fullflexible]
{prompt}
text for the task: {input_text}
Final answer should be at the end of your answer and its format should be like "Final answer: your_answer".
Generate output following the task description above.
Output:
Let's first understand the problem and devise a plan to solve the problem. Then, let's carry out the plan and solve the problem step by step.
\end{lstlisting}
\end{tcolorbox}

\section{Human-written Pseudocode Prompts}
\label{ssec:human-written}

\subsection{Human-written \codeprompt of Dyck Languages}
\begin{tcolorbox}[breakable, toprule at break=0pt, bottomrule at break=0pt,colback=white]
\begin{lstlisting}[style=prompt, columns=fullflexible]
def complete_dyck_languages(input_text):
    # Step 1: Initialize a stack to keep track of open parentheses and split the input text to identify and define all types of open parentheses in the text.
    stack = []
    character_list = input_text.split()
    open_to_close_parenthesis_dict = {"(": ")", "<": ">", "{": "}", "[": "]"}
    opening_parenthesis = ["(", "<", "{", "["]
    print(f"Parse characters in the input and initialize a stack to track of open parentheses. \nCurrent stack: {stack}. Parsed characters: {character_list}") 
    
    
    # Step 2: Through iteration over the input characters, identify opening parentheses among the input characters and add them to the stack.
    print("Check if a character is an opening parenthesis while iterating over the input characters.")
    for char in character_list:
        if char in opening_parenthesis:
		        print(f"Iteration {i+1}: Current character {char} is an opening parenthesis.")
            stack.append(char)
            print(f"Thus, we append {char} to the stack. Current stack after insertion: {', '.join(stack)}")
        
        # Step 3: For each open parentheses, find the corresponding closing parentheses and close the open parentheses.
        else:
            print(f"Iteration {i+1}: Current character {char} is not an opening parenthesis.\n Thus we delete the last item {stack[-1]} from the stack\n current stack before deletion: {" ".join(stack)} -> updated stack after deletion: {' '.join(stack[:-1]) if stack else 'empty'}")
            stack.pop() # Remove the last added open parentheses assuming a correct match.
    
    # Step 4: Generate the sequence of closing parentheses based on remaining open parentheses in the stack. 
    print(f"The resulting stack is {' '.join(stack)}.")
    print(f"We will need to pop out {' '.join(stack[::-1])} one by one in that order.")
    closing_list = [parentheses_pairs[opening] for opening in stack[::-1]]
    
    # Step 5: Output the completed sequence. Generate the input sequence concatenated with the generated closing sequence of parentheses, ensuring a well-formed structure.
    return " ".join(closing_list) 
\end{lstlisting}
\end{tcolorbox}

\subsection{Human-written \codeprompt of Geometric Shapes}
\begin{tcolorbox}[breakable, toprule at break=0pt, bottomrule at break=0pt,colback=white]
\begin{lstlisting}[style=prompt, columns=fullflexible]
def recognize_shape_from_svg(input_text):
    # Step 1: Get the SVG path data from the input text and generate the extracted SVG path.
    paths = parse_path(input_text)
    print("SVG paths:\n ", paths)
		
    # Step 2: Initialize a coordinate map that maps each coordinate with the other connected coordinates and the connection type.
    coordinate_map = dict()

    # Step 3: Update the coordinate map referring to the each SVG path.
    for i, path in enumerate(paths):
      coordinate_map = update_coordinate_map(coordinate_map, path)
      print(f"Step {i} - path: {path}, updated coordinate map: {coordinate_map}")

    # Step 4: Conduct calculation to analyze each characteristic of the shape.
    analysis_results_dict = analyze_characteristics(coordinate_map)
    print(f"Anlysis results: {analysis_results_dict}")

    # Step 5: Identify a geometric shape with reasons using the completed coordinates map and the analysis results.
    reason_for_the_decision, name_of_the_shape = identify_shape_with_explanation(coordinate_map, analysis_results_dict)
    print(f"Reason for the decision: {reason_for_the_decision}")
    print(f"Thus, the shape of the path is {name_of_the_shape}.")

    # Step 6: Find the corresponding option from the given options and only output the label of the option as the final answer to the question.
    options = parse_options(input_text)
    print(f"Options: {options}")
    answer = None
    for option in options:
      if name_of_the_shape in option:
        answer = option[:3]
    
    return answer
\end{lstlisting}
\end{tcolorbox}

% \begin{figure}
%     \begin{small}
%         \captionsetup{justification=centering, labelfont=bf, font=small}
%         \begin{minted}[fontsize=\footnotesize, frame=lines, framesep=2mm, baselinestretch=1.2, breaklines, breaksymbolleft={}, breaksymbolright={},bgcolor=Box2Color]{text}
% def ends_up_at_start(input_text):
%     # Step 1: Initialize coordinates and direction by setting the starting point at (0, 0) and face north.
%     cur_x, cur_y = 0, 0
%     cur_direction = 0

%     # Step 2: Identify and list up instructions from the input text.
%     instructions = parse_instructions(input_text)
    
%     # Step 3: Process each instruction and update the current coordinates and direction. In order to keep track of changes, output the instruction, current and updated coordinates and direction.
%     for i, instruction in enumerate(instructions):
%         new_x, new_y, new_direction = process_instruction(instruction, cur_x, cur_y, cur_direction) # process instruction to calculate new position and direction
%         print(f"Step {i}: {instruction} - current coordinates: ({cur_x}, {cur_y}), current direction: {cur_direction} -> updated coordinates: ({new_x}, {new_y}), updated direction: {new_direction}")
%         cur_x, cur_y, cur_direction = new_x, new_y, new_direction

%     # Step 4: Return "yes" if the final coordinates are (0, 0). Otherwise, return "no" as the final answer.
%     return 'yes' if cur_x == 0 and cur_y == 0 else 'no'
%         \end{minted}
%     \end{small}
%     \caption{Human-written \codeprompt of Navigate}
%     \label{fig:human_written_prompt:naviate}
% \end{figure}

\subsection{Human-written \codeprompt of Navigate}
\begin{tcolorbox}[breakable, toprule at break=0pt, bottomrule at break=0pt,colback=white]
\begin{lstlisting}[style=prompt, columns=fullflexible]
def ends_up_at_start(input_text):
    # Step 1: Initialize coordinates and direction by setting the starting point at (0, 0) and face north.
    cur_x, cur_y = 0, 0
    cur_direction = 0

    # Step 2: Identify and list up instructions from the input text.
    instructions = parse_instructions(input_text)
    
    # Step 3: Process each instruction and update the current coordinates and direction. In order to keep track of changes, output the instruction, current and updated coordinates and direction.
    for i, instruction in enumerate(instructions):
        new_x, new_y, new_direction = process_instruction(instruction, cur_x, cur_y, cur_direction) # process instruction to calculate new position and direction
        print(f"Step {i}: {instruction} - current coordinates: ({cur_x}, {cur_y}), current direction: {cur_direction} -> updated coordinates: ({new_x}, {new_y}), updated direction: {new_direction}")
        cur_x, cur_y, cur_direction = new_x, new_y, new_direction

    # Step 4: Return "yes" if the final coordinates are (0, 0). Otherwise, return "no" as the final answer.
    return 'yes' if cur_x == 0 and cur_y == 0 else 'no'
\end{lstlisting}
\end{tcolorbox}

\subsection{Human-written \codeprompt of Reasoning about Colored Objects}
\begin{tcolorbox}[breakable, toprule at break=0pt, bottomrule at break=0pt,colback=white]
\begin{lstlisting}[style=prompt, columns=fullflexible]
def solve_colored_objects(input_text):
    # Step 1: Start by identifying the objects along with their associated properties, such as color and spatial positioning from the input text. Show the list of objects.
    objects_list = extract_objects(input_text)
    print("Objects and their properties:", objects_list)

    # Step 2: Identify the specific question asked. Determine whether the question is about identifying the color of a specific object, counting objects of a certain color, or reasoning about the spatial arrangement of objects and output the question type.
    question = extract_question(input_text)
    print("Question specifics:", question)

    # Step 3: Identify and list up available options provided in the input text.
    options = input_text.split("\n")[-5:]

    # Step 4: Process according to the question type and show what the question type is:
    # If the question is about identifying color, identify and ouput the target object the question is asking for the color of. Determine and output its color. 
    if question['type'] == 'identify_color':
        print("Question type is = identify_color")
        print(f"Identifying color for: {question['details']}")
        target_object = target(objects_list, question['details'])
        print(f"The question is asking for the color of : {target_object}")
        pre_answer = extract_color(target_object, question['details'])
        print(f"Identified color: {pre_answer}")

    # If the question is about counting objects, identify and ouput the objects the question is asking for the number of. Go through each object in the list in steps and count each object. Show the counting steps. Output the final number of objects that meet the specified criteria (e.g., a specific color). 
    elif question['type'] == 'count_objects':
        print("Question type is = count_objects")
        print(f"Counting objects for: {question['details']}")
        print("Total iterations:", len(objects_list))
        for i, object in enumerate(objects_list):
            single_object_count = count_single_object(object, question['details'])
            intermediate_count += single_object_count
            print(f"Step ({i}) - {object}: {single_object_count}, Intermediate count: {intermediate_count}")
        pre_answer = count_objects(objects_list, question['details'])
        print(f"Objects count: {pre_answer}")

    # If the question is about spatial reasoning, identify and ouput the relative positions the question is asking for. Arrange the objects from left to right and output the order. Determine the relative positions of objects and output the result.
    elif question['type'] == 'spatial_reasoning':
        print("Question type is = spatial_reasoning")
        print(f"Applying spatial reasoning for: {question['details']}")
        arranged_object = arrange_from_left_to_right(objects_list)
        print(f"Arraged objects: {arranged_object})
        pre_answer = spatial_reasoning(arranged_object, question['details'])
        print(f"Spatial reasoning result: {pre_answer}")

    # Step 5: Recall the identified options and match the outcome of Step 4 (the identified color, the count of objects, or the result of spatial reasoning) with the provided options to determine the correct answer.
    answer = find_correct_option(pre_answer, options)
    
    # Step 6: Return the final answer chosen at Step 5.
    return answer\end{lstlisting}
\end{tcolorbox}

\subsection{Human-written \codeprompt of Temporal Sequences}
\begin{tcolorbox}[breakable, toprule at break=0pt, bottomrule at break=0pt,colback=white]
\begin{lstlisting}[style=prompt, columns=fullflexible]

def solve_temporal_sequences_quiz(input_text):
    # Step 1: Identify statements and options from the input_text and output the statements.
    statement_text, option_text = input_text.split("\nOptions:\n")
    parts = statement_text.split("\n")
    statements = parts[1:-2]
    options = option_text.split("\n")
    print("Statements:", statements)

    # Step 2: Check the start and end of the possible time.
    print("Start of the possible time: ", parts[0])
    print("End of the possible time: ", parts[-2])
    
    # Step 3: Initialize an available time map with the time slots in the options and output it. The time slots are marked as 'free' initially.
    available_time_map = {option[4:]: "free" for option in options}
    print(f"Initial available time dictionary: {available_time_map}")

    # Step 4: Sequentially go through each statement, marking the times when the individual was seen or known to be engaged in specific activities. In this step, you should generate the target time slots and the updated available time map according to the statement.
    for i, statement in enumerate(statements):
        event, time_span = extract_information(statement)
        print(f"\nStep {i}: {statement}")
        print(f"current time occupation: {available_time_map}")
        print(f"Time span to be occupied: {time_span}")
        available_time_map[time_span] = "not available"
        print(f"updated time occupation: {available_time_map}")

    # Step 5: By checking the available time map, identify which time slot is marked as 'free'. For each time slot, output the time slot is free or not available.
    for key in available_time_map:
        if available_time_map[key] == "free":
            print(f"{key} is free.")
            free_time = key
        else:
            print(f"{key} is not available.")
    # Step 6: Review the provided options and return the one that matches the identified free time slot in Step 5.
    print(f"Options:\n{option_text}")
    for option in options:
        if free_time in option:
            return option
\end{lstlisting}
\end{tcolorbox}

\subsection{Human-written \codeprompt of Tracking Shuffled Objectives}
\begin{tcolorbox}[breakable, toprule at break=0pt, bottomrule at break=0pt,colback=white]
\begin{lstlisting}[style=prompt, columns=fullflexible]
def track_swaps(input_text):
    # Step 1: Identify Initial State. Begin by identifying and outputing the initial state of all objectives (e.g., who holds which ball or who is dancing with whom) from the input text before any swaps happen.
    state_dict = find_initial_state(input_text)
    print(f"Initial state: {state_dict}")

    # Step 2: Identify and output the sequences of swaps from the input text. Each swap should be understood in terms of who exchanges with whom.
    swap_sequences_list = find_swap_sequences(input_text)
    print("Swap sequences: ", swap_sequences_list)
    print("Total iterations: ", len(swap_sequences_list))

    # Step 3: Carry out the swaps. For each swap in swap sequences, sequentially update and output the current status of objectives by exchanging them between the two participants involved in the swap.
    for i, sequence in enumerate(swap_sequences_list):
        player1, player2 = extract_player(sequence)
        state_dict[player1], state_dict[player2] = state_dict[player2], state_dict[player1]
        print(f"({i}) {sequence} -> {state_dict}")

    Step 4: Understand the Question. After processing all swaps, identify what the question is asking for in the input text and output the question.
    question = extract_question(input_text)
    print("Question:", question)

    Step 5: Analyze Options. Examine and output the provided options in the input text.
    options = input_text.split("\n")[-5:]
    print("Options:", options)

    Step 6: Determine the Correct Option. Using the updated state after all swaps, determine which option correctly answers the question and output the answer.
    answer = find_correct_option(question, options, state_dict)

    return answer
\end{lstlisting}
\end{tcolorbox}

\subsection{Human-written \codeprompt of Web of Lies}
\begin{tcolorbox}[breakable, toprule at break=0pt, bottomrule at break=0pt,colback=white]
\begin{lstlisting}[style=prompt, columns=fullflexible]

def evaluate_boolean_word_problem(input_text):
    # Step 1: Divide the input text into individual statements and the final question. Output each statements.
    statements = input_text.split("")[:-1]
    question = input_text.split("")[-1]
    print("Parsed statements:", statements)
    
    # Step 2: Create a Truth Map to keep track of the assumed truthfulness of each person mentioned in the statements. No truth values are assigned initially.
    truth_map = {statement.split()[0]: None for statement in statements}

    # Step 3: Analyze Each Statement. For each statement, first output the statement number and the statement. identify the subject person (who makes the statement), the object person (who the statement is about), and the expected truth value (whether the object person is said to tell the truth or lie). Output the current statement under analysis along with the object person and the expected truth value for clarity.
    for i, statement in enumerate(statements):
        print(f"({i}): {statement}")
        speaker, target_person, expected_truth_value_of_target_person = extract_person_and_truth_value(statement) # speaker - says - target_person - expected_truth_value_of_target_person

        print(f"{speaker} says : {target_person} - {expected_truth_value_of_target_person}")
        print(f"Truth value of {target_person}: {truth_map[target_person]}")

        # Step 4: Update the Truth Map based on the analysis of each statement. If the statement's claim aligns with the current assumption about the object person's truthfulness, mark the subject person as truthful. Otherwise, mark them as untruthful. After each update, print the name of the person being updated, their determined truth value, and the updated truth map to track changes.
        if truth_map[target_person] == None: # if the statement does not need to be checked
            print(f"{expected_truth_value_of_target_person} matches {truth_map[target_person]}")
            truth_map[target_person] = expected_truth_value_of_target_person
        else:
            print(f"{expected_truth_value_of_target_person} does not match {truth_map[target_person]}")
            if truth_map[target_person] == expected_truth_value_of_target_person: # the speaker tells the truth
                truth_map[speaker] = True
            else: # the speaker lies
                truth_map[speaker] = False

        print(f"Person to update: {speaker} - {truth_map[speaker]}")

        print(f"updated truth_map: {truth_map}")
        print("\n\n")
    
    # Step 5: Using the completed truth map, determine the truthfulness of the person in the final question. Output the question for reference before making the determination.
    print("question:", question)
    target_person_in_question = extract_target_person_from_question(question)
    target_truth_value = truth_map[target_person_in_question]
    print(f"{target_person_in_question}: {target_truth_value})

    # Step 6: Based on the evaluation, output "Yes" if the person in the final question is determined to tell the truth, otherwise output "No".
    answer = "Yes" if target_truth_value else "No"

    return answer

\end{lstlisting}
\end{tcolorbox}

\section{Generated Analyses}

\subsection{Generated Analysis of Dyck Languages}
\begin{tcolorbox}[breakable, toprule at break=0pt, bottomrule at break=0pt,colback=white]
\begin{lstlisting}[style=text, columns=fullflexible]
Explanation:
This task involves completing a sequence of parentheses to ensure they are closed properly. The challenge is to understand the nesting structure of the parentheses and add the appropriate closing brackets to match each opening bracket.

Analyzing the Task:
1. Parentheses Matching: Each opening parenthesis should have a corresponding closing parenthesis to ensure the sequence is valid.

2. Nested Parentheses: The parentheses can be nested within each other, so it's crucial to maintain the correct nesting structure.

3. Proper Closure: The goal is to add the necessary closing brackets to complete the sequence and make sure all parentheses are closed properly.

Constructing the Code Prompt:
1. Parsing Input: Extract the sequence of parentheses from the input text.

2. Stack Implementation: Use a stack data structure to keep track of the opening parentheses that need closing.

3. Iterating Over the Sequence: Iterate over each character in the sequence and push opening parentheses onto the stack.

4. [Important] Closing Parentheses: When encountering a closing parenthesis, pop the corresponding opening parenthesis from the stack to ensure proper closure.

5. Generating Output: After processing the entire sequence, the stack should be empty if all parentheses are closed properly. If not, add the necessary closing brackets to match the remaining opening brackets.

6. Returning the Result: Output the generated closing brackets as a string, separated by spaces, to complete the sequence.
\end{lstlisting}
\end{tcolorbox}

\subsection{Generated Analysis of Geometric Shapes}
\begin{tcolorbox}[breakable, toprule at break=0pt, bottomrule at break=0pt,colback=white]
\begin{lstlisting}[style=text, columns=fullflexible]
Explanation:
This task involves analyzing SVG path elements to determine the shape that is being drawn. The path elements consist of a series of commands that define the movements and curves in the SVG image. By understanding these commands and their implications, we can identify the shape being drawn.

Analyzing the Task:
1. SVG Path Commands: The SVG path element contains commands like M (move to), L (line to), A (arc), etc., each of which contributes to the overall shape being drawn.

2. Interpreting Path Commands: Each command in the path element corresponds to a specific action, such as moving to a point, drawing a line, or creating an arc. Understanding these commands is crucial to identifying the shape.

3. Recognizing Geometric Shapes: Different combinations of path commands result in different geometric shapes like circles, triangles, rectangles, etc. We need to match the commands to the shapes they represent.

Constructing the Code Prompt:
1. Parsing SVG Path Element: Extract the path commands from the SVG path element provided in the task. This involves splitting the string and identifying each command.

2. Processing Path Commands: Iterate over each command in the path element and execute the corresponding action. This involves:
- Moving to a new point (M command).
- Drawing a line to a point (L command).
- Drawing an arc (A command).
- Other commands as necessary.

3. [Important] Tracking Coordinates: Keep track of the coordinates as the path commands are executed. This involves updating the current position based on the commands.

4. Determining the Shape: After processing all commands, analyze the resulting path to determine the shape being drawn. This can be done by comparing the final path with the characteristics of known shapes.

5. Matching with Provided Options: Compare the identified shape with the options provided in the task to select the correct answer.

6. Returning the Result: Return the identified shape as the output in the specified format ('(A)', '(B)', '(C)', ...).

By following these steps and accurately interpreting the SVG path commands, we can determine the shape being drawn and select the correct option from the given choices.
\end{lstlisting}
\end{tcolorbox}

\subsection{Generated Analysis of Navigate}
\begin{tcolorbox}[breakable, toprule at break=0pt, bottomrule at break=0pt,colback=white]
\begin{lstlisting}[style=text, columns=fullflexible]
Explanation:
This task involves following a series of instructions related to movement and direction to determine if the final position is the same as the starting point. The challenge lies in accurately tracking the movements and rotations to deduce the final position.

Analyzing the Task:
1. Movement Tracking: Keep track of the steps taken in each direction (forward, backward, left, right) to determine the final position.

2. Directional Changes: Account for any rotations (turning left or right) that may alter the orientation during movement.

3. Spatial Reasoning: Apply logical reasoning to calculate the final position based on the cumulative effect of the movements and rotations.

Constructing the Code Prompt:
1. Extracting Instructions: Parse the input text to extract the sequence of movements and rotations.

2. Processing Movements:
   - Initialize variables to track the current position (x, y coordinates) and orientation (direction faced).
   - Iterate over each instruction, updating the position and orientation accordingly.

3. [Important] Determining the Final Position:
   - Calculate the final position based on the cumulative effect of movements and rotations.
   - Check if the final position matches the starting point to determine if the participant returns to the starting point.

4. Matching with Options: Compare the final position with the starting point to determine if the participant returns to the starting point.

5. Returning the Final Answer: Output 'Yes' if the participant returns to the starting point, 'No' otherwise.
\end{lstlisting}
\end{tcolorbox}

\subsection{Generated Analysis for Reasoning about Colored Objects}
\begin{tcolorbox}[breakable, toprule at break=0pt, bottomrule at break=0pt,colback=white]
\begin{lstlisting}[style=text, columns=fullflexible]
Explanation:
This task involves identifying specific attributes of items based on their descriptions and positions relative to each other. The task requires careful reading of the descriptions and analyzing the spatial relationships between the items.

Analyzing the Task:
The task involves interpreting the descriptions of items and their spatial relationships to answer specific questions. This task does not require variable tracking but rather a free format reasoning approach to extract the necessary information and make logical deductions.

Constructing the Code Prompt:

1. Parse the descriptions: Extract the descriptions of items and their colors from the input text.

2. Identify the spatial relationships: Determine the positions of items relative to each other based on the descriptions provided.

3. [Free format reasoning] Derive the answer with explanation: Analyze the descriptions and spatial relationships to answer the specific question posed in the task. Use a function like 'derive_answer_with_explanation' to provide both the reasoning and the correct answer.

4. Match the answer with the options: Compare the derived answer with the given options and select the correct one.

5. Return the final answer: Output the final answer in the specified format.

By following these steps, the task can be effectively solved by reasoning through the descriptions and spatial relationships of the items provided in the task instances.
\end{lstlisting}
\end{tcolorbox}

\subsection{Generated Analysis of Temporal Sequences}
\begin{tcolorbox}[breakable, toprule at break=0pt, bottomrule at break=0pt,colback=white]
\begin{lstlisting}[style=text, columns=fullflexible]
Explanation:
This task involves determining the possible time intervals during which a person could have visited a specific location based on the given schedule of events throughout the day. The goal is to identify the time frames that are not conflicting with other activities.

Analyzing the Task:
1. Establishing the Timeline: Understanding the sequence of events and the times at which they occurred is crucial for determining the available time slots.

2. Identifying Conflicting Activities: Recognizing the time intervals during which the person was engaged in other activities that would prevent them from visiting the specified location.

3. Determining Possible Visitation Times: By eliminating conflicting time intervals, the remaining time slots represent the possible visitation times.

Constructing the Code Prompt:
1. Extracting Event Information: Parse the given schedule of events to extract the times at which each activity occurred. This may involve creating a data structure to store this information.

2. Identifying Conflicting Time Intervals: Determine the time intervals during which the person was occupied with other activities that would prevent them from visiting the specified location.

3. [Important] Calculating Available Time Slots: By subtracting the conflicting time intervals from the total day duration, identify the time frames during which the person could have visited the specified location.

4. Matching with Options: Compare the possible visitation times with the provided options to determine the correct answer.

5. Returning the Correct Option: Return the letter corresponding to the time interval that matches the calculated possible visitation times.

This task involves a logical deduction process based on the given schedule of events to determine the feasible time intervals for visiting the specified location. By systematically analyzing the timeline and eliminating conflicting activities, the correct answer can be identified. The solution should provide a clear and accurate assessment of the available time slots for the visitation.
\end{lstlisting}
\end{tcolorbox}

\subsection{Generated Analysis of Tracking Shuffled Objectives}
\begin{tcolorbox}[breakable, toprule at break=0pt, bottomrule at break=0pt,colback=white]
\begin{lstlisting}[style=text, columns=fullflexible]
Explanation:
This task involves tracking the positions or gifts of individuals as they swap with each other in a sequence of exchanges. The challenge is to keep track of the swaps accurately and determine the final position or gift of a specific individual at the end of the sequence.

Analyzing the Task:
1. Initial Assignments: Understanding the starting positions or gifts of each individual is crucial for tracking the swaps accurately.

2. Pairwise Swaps: Identifying the pairs of individuals who swap positions or gifts at each step in the sequence.

3. Tracking Changes: Keeping a record of the swaps and updating the positions or gifts of individuals accordingly.

Constructing the Code Prompt:
1. Initialize Positions or Gifts: Set up a dictionary mapping each individual to their initial position or gift.

2. Define Swap Pairs: Create a list of tuples representing the pairs of individuals who swap positions or gifts at each step.

3. [Important] Process Swaps: Iterate over the swap pairs, update the positions or gifts of the individuals involved in each swap. Use an if-else statement to handle different swap scenarios.

4. Determine the Final Position or Gift: After processing all swaps, identify the final position or gift of the specified individual.

5. Match and Output the Answer: Parse the options from the input text, find the corresponding option from the given options, and only output the label of the option as the final answer to the question.
\end{lstlisting}
\end{tcolorbox}

\subsection{Generated Analysis of Web of Lies}
\begin{tcolorbox}[breakable, toprule at break=0pt, bottomrule at break=0pt,colback=white]
\begin{lstlisting}[style=text, columns=fullflexible]
Explanation:
This task involves determining the truthfulness of a statement made by one individual based on the statements made by others in a chain. The task requires understanding the relationships between truth-tellers and liars and applying logical reasoning to determine the final answer.

Analyzing the Task:
1. Establishing Truth Relationships: Each person's statement about another person can be categorized as either true or false. This forms the basis of determining who tells the truth and who lies.

2. Propagating Truthfulness: By analyzing the statements in a sequential manner, the truthfulness of each person can be deduced based on the statements made by others.

3. Identifying the Final Question: The task usually asks whether a specific person tells the truth or not based on the chain of statements.

Constructing the Code Prompt:
1. Parsing Statements: Extract the statements made by each person from the input text. This involves identifying who is talking about whom and whether they are telling the truth or lying.

2. Establishing Truth Relationships: Create a dictionary to store the truthfulness of each person based on the statements made by others. This dictionary will be updated as the statements are processed.

3. [Important] Analyzing Statements: Iterate over each statement and update the truthfulness of the individuals involved based on the logic that if A says B lies, then A is telling the truth if B is a liar, and vice versa. This step involves logical reasoning and updating the truth dictionary.

4. Extracting the Final Question: Identify the specific question asked in the input text regarding the truthfulness of a particular person.

5. Determining the Answer: Based on the final truthfulness of the person in question as determined by the logic and the statements provided, select 'Yes' if the person tells the truth and 'No' if they do not.

By following these steps and applying logical reasoning to the statements provided, the code can accurately determine whether the specified individual tells the truth or not.
\end{lstlisting}
\end{tcolorbox}
\section{Generated Pseudocode Prompts}

\subsection{Generated \codeprompt of Dyck Languages}
\begin{tcolorbox}[breakable, toprule at break=0pt, bottomrule at break=0pt,colback=white]
\begin{lstlisting}[style=prompt, columns=fullflexible]
def complete_dyck_languages(input_text):
    # Step 1: Parse the input text to extract the sequence of parentheses.
    parentheses_sequence = extract_parentheses(input_text)
    print("Parentheses sequence:", parentheses_sequence)

    # Step 2: Initialize a stack to keep track of opening parentheses that need closing.
    stack = []

    # Step 3: Iterate over each character in the sequence to handle opening and closing parentheses.
    for i, char in enumerate(parentheses_sequence):
        if char in ['(', '[', '{', '<']:
            stack.append(char)
            print(f"Step ({i}) - Pushed {char} to stack")
        elif char in [')', ']', '}', '>']:
            if len(stack) == 0:
                print(f"Step ({i}) - No matching opening bracket for {char}")
            else:
                opening_bracket = stack.pop()
                print(f"Step ({i}) - Matched {opening_bracket} with {char}")
    
    # Step 4: Check if the stack is empty after processing the entire sequence.
    if len(stack) > 0:
        print("Remaining unmatched brackets in stack:", stack)
        # Step 5: Generate the necessary closing brackets to match the remaining opening brackets.
        closing_brackets = generate_closing_brackets(stack)
        print("Generated closing brackets:", closing_brackets)
    else:
        print("All brackets are properly matched.")

    # Step 6: Return the generated closing brackets as the final answer.
    return closing_brackets
\end{lstlisting}
\end{tcolorbox}

\subsection{Generated \codeprompt of Geometric Shapes}
\begin{tcolorbox}[breakable, toprule at break=0pt, bottomrule at break=0pt,colback=white]
\begin{lstlisting}[style=prompt, columns=fullflexible]
def recognize_shape_from_svg(svg_path_element):
    # Step 1: Parse the SVG path element to extract the path commands. Output the extracted path commands.
    path_commands = extract_path_commands(svg_path_element)
    print("Extracted path commands:", path_commands)

    # Step 2: Process each path command to execute the corresponding action. Keep track of the coordinates as the commands are executed.
    current_position = (0, 0)
    for command in path_commands:
        action, coordinates = process_path_command(command, current_position)
        print(f"Executing command: {command} - Action: {action} - New coordinates: {coordinates}")
        current_position = coordinates

    # Step 3: Determine the shape based on the processed path commands. This involves analyzing the final path drawn.
    identified_shape = identify_shape(path_commands)
    print("Identified shape:", identified_shape)

    # Step 4: Match the identified shape with the provided options to select the correct answer.
    options = extract_options(svg_path_element)
    answer = match_shape_with_options(identified_shape, options)

    # Step 5: Return the identified shape as the final answer.
    return answer
\end{lstlisting}
\end{tcolorbox}
% \begin{figure}
%     \begin{small}
%         \captionsetup{justification=centering, labelfont=bf, font=small}
%         \begin{minted}[fontsize=\footnotesize, frame=lines, framesep=2mm, baselinestretch=1.2, breaklines, breaksymbolleft={}, breaksymbolright={},bgcolor=Box2Color]{text}
% def ends_up_at_start(input_text):
%     # Step 1: Extract the sequence of movements and rotations from the input text.
%     instructions = extract_instructions(input_text)
%     print("Instructions:", instructions)

%     # Step 2: Initialize variables to track the current position and orientation.
%     x, y = 0, 0
%     orientation = "N"
%     print(f"Initial position: ({x}, {y}), Orientation: {orientation}")

%     # Step 3: Process each instruction to update the position and orientation accordingly.
%     for instruction in instructions:
%         x, y, orientation = process_instruction(instruction, x, y, orientation)
%         print(f"Instruction: {instruction} -> Position: ({x}, {y}), Orientation: {orientation}")

%     # Step 4: Determine the final position after following all instructions.
%     final_position = (x, y)
%     print("Final Position:", final_position)

%     # Step 5: Check if the final position matches the starting point to determine if the participant returns to the starting point.
%     if final_position == (0, 0):
%         return 'Yes'
%     else:
%         return 'No'
%         \end{minted}
%     \end{small}
%     \caption{Generated \codeprompt of Navigate}
%     \label{fig:generated_prompt:naviate}
% \end{figure}

\subsection{Generated \codeprompt of Navigate}
\begin{tcolorbox}[breakable, toprule at break=0pt, bottomrule at break=0pt,colback=white]
\begin{lstlisting}[style=prompt, columns=fullflexible]
def ends_up_at_start(input_text):
    # Step 1: Extract the sequence of movements and rotations from the input text.
    instructions = extract_instructions(input_text)
    print("Instructions:", instructions)

    # Step 2: Initialize variables to track the current position and orientation.
    x, y = 0, 0
    orientation = "N"
    print(f"Initial position: ({x}, {y}), Orientation: {orientation}")

    # Step 3: Process each instruction to update the position and orientation accordingly.
    for instruction in instructions:
        x, y, orientation = process_instruction(instruction, x, y, orientation)
        print(f"Instruction: {instruction} -> Position: ({x}, {y}), Orientation: {orientation}")

    # Step 4: Determine the final position after following all instructions.
    final_position = (x, y)
    print("Final Position:", final_position)

    # Step 5: Check if the final position matches the starting point to determine if the participant returns to the starting point.
    if final_position == (0, 0):
        return 'Yes'
    else:
        return 'No'
\end{lstlisting}
\end{tcolorbox}

\subsection{Generated \codeprompt for Reasoning about Colored Objects}
\begin{tcolorbox}[breakable, toprule at break=0pt, bottomrule at break=0pt,colback=white]
\begin{lstlisting}[style=prompt, columns=fullflexible]
def solve_colored_objects(input_text):
    # Step 1: Extract the descriptions of items and their colors from the input text.
    items = parse_items(input_text)
    print("Items on the surface:\n", items)

    # Step 2: Determine the positions of items relative to each other based on the descriptions provided.
    spatial_relationships = analyze_spatial_relationships(items)
    print("Spatial relationships between items:\n", spatial_relationships)

    # Step 3: Derive the answer with explanation by analyzing the descriptions and spatial relationships.
    question = identify_question(input_text)
    print("The question is:", question)
    reason, answer = derive_answer_with_explanation(items, spatial_relationships, question)
    print("Reasoning for the answer:", reason)

    # Step 4: Compare the derived answer with the given options and select the correct one.
    options = extract_options(input_text)
    print("Answer options:\n", options)
    final_answer = None
    for option in options:
        if answer in option:
            final_answer = option[:3]
            break

    # Step 5: Return the final answer.
    return final_answer
\end{lstlisting}
\end{tcolorbox}

\subsection{Generated \codeprompt of Temporal Sequences}
\begin{tcolorbox}[breakable, toprule at break=0pt, bottomrule at break=0pt,colback=white]
\begin{lstlisting}[style=prompt, columns=fullflexible]
def solve_temporal_sequences_quiz(input_text):
    # Step 1: Extract event information from the input text to understand the timeline of activities throughout the day.
    events_list = extract_events(input_text)
    print("Events and their timings:", events_list)

    # Step 2: Identify conflicting time intervals when the person was engaged in other activities that would prevent them from visiting the specified location.
    conflicting_intervals = find_conflicting_intervals(events_list)
    print("Conflicting time intervals:", conflicting_intervals)

    # Step 3: Calculate the available time slots by subtracting conflicting intervals from the total day duration.
    available_intervals = calculate_available_intervals(conflicting_intervals)
    print("Available time intervals for visitation:", available_intervals)

    # Step 4: Match the available time slots with the provided options to determine the correct answer.
    options = input_text.split("\n")[-5:]

    # Step 5: Return the correct option corresponding to the time interval that aligns with the calculated possible visitation times.
    answer = find_correct_option(available_intervals, options)

    return answer
\end{lstlisting}
\end{tcolorbox}

\subsection{Generated \codeprompt of Tracking Shuffled Objectives}
\begin{tcolorbox}[breakable, toprule at break=0pt, bottomrule at break=0pt,colback=white]
\begin{lstlisting}[style=prompt, columns=fullflexible]
def track_swaps(input_text):
    # Step 1: Initialize Positions or Gifts
    initial_assignments = extract_initial_assignments(input_text)
    print("Initial assignments:", initial_assignments)

    # Step 2: Define Swap Pairs
    swap_pairs = extract_swap_pairs(input_text)
    print("Swap pairs:", swap_pairs)

    # Step 3: Process Swaps
    final_assignments = initial_assignments.copy()
    for pair in swap_pairs:
        if pair[0] in final_assignments and pair[1] in final_assignments:
            temp = final_assignments[pair[0]]
            final_assignments[pair[0]] = final_assignments[pair[1]]
            final_assignments[pair[1]] = temp
        print(f"After swapping {pair[0]} and {pair[1]}: {final_assignments}")

    # Step 4: Determine the Final Position or Gift
    final_person = identify_final_person(input_text)
    final_position = final_assignments[final_person]
    print(f"The final position or gift of {final_person} is: {final_position}")

    # Step 5: Match and Output the Answer
    options = extract_options(input_text)
    print("Options:", options)
    answer = None
    for option in options:
        if final_position in option:
            answer = option[:3]
    
    return answer
\end{lstlisting}
\end{tcolorbox}
% \begin{figure}
%     \begin{small}
%         \captionsetup{justification=centering, labelfont=bf, font=small}
%         \begin{minted}[fontsize=\footnotesize, frame=lines, framesep=2mm, baselinestretch=1.2, breaklines, breaksymbolleft={}, breaksymbolright={},bgcolor=Box2Color]{text}
% def evaluate_boolean_word_problem(input_text):
%     # Step 1: Identify the statements and the question from the input text.
%     statements, question = input_text.split("?")
%     statements = statements.split(". ")
%     print("Statements:", statements)
%     print("Question:", question)

%     # Step 2: Initialize a truth dictionary to keep track of who tells the truth and who lies.
%     truth_dict = {}

%     # Step 3: Process each statement to update the truth dictionary based on the logic provided.
%     for statement in statements:
%         person1, action, person2 = statement.split(" ")
%         if action == "lies":
%             truth_dict[person1] = not truth_dict[person2]
%         else:
%             truth_dict[person1] = truth_dict[person2]
%         print(f"{person1} says {person2} {action}. {person1} tells the truth: {truth_dict[person1]}")

%     # Step 4: Determine the truthfulness of the person in question based on the truth dictionary.
%     person_to_check = question.split(" ")[-2]
%     answer = 'Yes' if truth_dict[person_to_check] else 'No'
    
%     return answer
%         \end{minted}
%     \end{small}
%     \caption{Generated \codeprompt of Web of Lies}
%     \label{fig:generated_prompt:web_of_lies}
% \end{figure}

\subsection{Generated \codeprompt of Web of Lies}
\begin{tcolorbox}[breakable, toprule at break=0pt, bottomrule at break=0pt,colback=white]
\begin{lstlisting}[style=prompt, columns=fullflexible]
def evaluate_boolean_word_problem(input_text):
    # Step 1: Identify the statements and the question from the input text.
    statements, question = input_text.split("?")
    statements = statements.split(". ")
    print("Statements:", statements)
    print("Question:", question)

    # Step 2: Initialize a truth dictionary to keep track of who tells the truth and who lies.
    truth_dict = {}

    # Step 3: Process each statement to update the truth dictionary based on the logic provided.
    for statement in statements:
        person1, action, person2 = statement.split(" ")
        if action == "lies":
            truth_dict[person1] = not truth_dict[person2]
        else:
            truth_dict[person1] = truth_dict[person2]
        print(f"{person1} says {person2} {action}. {person1} tells the truth: {truth_dict[person1]}")

    # Step 4: Determine the truthfulness of the person in question based on the truth dictionary.
    person_to_check = question.split(" ")[-2]
    answer = 'Yes' if truth_dict[person_to_check] else 'No'
    
    return answer
\end{lstlisting}
\end{tcolorbox}
\section{Qualitative Analysis}
\label{ssec:qual_analysis}
We conduct a qualitative analysis by comparing the outputs of our approach (\ours) with those of the baseline methods. This comparison is presented across Tables\ref{tab:dyck_comparsion},\ref{tab:geometric_comparision},\ref{tab:navigate_comparison},\ref{tab:reasoning_comparison},\ref{tab:temporal_comparison},\ref{tab:tracking_comparison}, and \ref{tab:web_comparison}.

\begin{table}[ht]
\small
\begin{tabularx}{\textwidth}{lX}
\toprule
Example Instance & Complete the rest of the sequence, making sure that the parentheses are closed properly. Input: ( \{ \{ \} \} \\
\midrule
Ground truth Answer & \textbf{)} \\
\\
\midrule
\textbf{Method} & \textbf{Output} \\
\midrule
\midrule
Direct Prompting & 
\begin{tabular}[t]{@{}l@{}}
Final answer: \{ \} ) \textbf{(Wrong)}
\end{tabular} \\
\midrule
Zero-shot Prompting & 
\begin{tabular}[t]{@{}l@{}}
Step 1: ( \{ \{ \} \} ) \\
Step 2: ( \{ \{ \} \} ) \\
Therefore, the answer is \} \textbf{(Wrong)}
\end{tabular} \\
\midrule
PoT & 
\begin{tabular}[t]{@{}l@{}}
def solution(): \\
\hspace*{1em}stack = [] \\
\hspace*{1em}sequence = "( \{ \{ \} \}" \\
\\
\hspace*{1em}for char in sequence: \\
\hspace*{2em}if char == "(" or char == "\{": \\
\hspace*{3em}stack.append(char) \\
\hspace*{2em}elif char == ")" and stack[-1] == "(": \\
\hspace*{3em}stack.pop() \\
\hspace*{2em}elif char == "\}" and stack[-1] == "\{": \\
\hspace*{3em}stack.pop() \\
\\
\hspace*{1em}for char in stack: \\
\hspace*{2em}if char == "(": \\
\hspace*{3em}sequence += ")" \\
\hspace*{2em}elif char == "\{": \\
\hspace*{3em}sequence += "\}" \\
\\
\hspace*{1em}return sequence \\
\\
solution() \\
\texttt{>>>} ( \{ \{ \} \}) \textbf{(Wrong)}
\end{tabular} \\
\midrule
NL Planning & 
\begin{tabular}[t]{@{}l@{}}
Step 1: Opening brackets in the input sequence: ( \{ \{ \\
Step 2: Corresponding closing brackets: \\
\texttt{( -> )} \\
\texttt{\{ -> \}} \\
\texttt{\{ -> \}} \\
Step 3: Closing brackets in order: ) \} \} \\
\\
Final answer: ) \} \} \textbf{(Wrong)}
\end{tabular} \\
\midrule
Ours & 
\begin{tabular}[t]{@{}l@{}}
Bracket sequence: ['(', '\{', '\{', '\}', '\}'] \\
Step 0 - bracket: ( \\
Updated stack: ['('] \\
Step 1 - bracket: \{ \\
Updated stack: ['(', '\{'] \\
Step 2 - bracket: \{ \\
Updated stack: ['(', '\{', '\{'] \\
Step 3 - bracket: \} \\
Updated stack: ['(', '\{'] \\
Step 4 - bracket: \} \\
Updated stack: ['('] \\
Rest of sequence: ) \\
Stack is not empty. The sequence is not properly balanced. \\
Final answer: ) \textbf{(Correct)}
\end{tabular} \\
\bottomrule
\end{tabularx}
\centering
\caption{A comparison of results for Dyck Languages between the baseline methods and \ours.}
\label{tab:dyck_comparsion}
\end{table}
\begin{table}[ht]
\small
\begin{tabularx}{\textwidth}{lX}
\toprule
Example Instance & This SVG path element \texttt{<}path d="M 38.00,62.00 L 48.00,60.00 L 51.00,49.00 L 54.00,60.00 L 65.00,62.00 L 54.00,64.00 L 51.00,74.00 L 48.00,64.00 L 38.00,62.00"\texttt{/>} draws a
Options:
(A) circle
(B) heptagon
(C) hexagon
(D) kite
(E) line
(F) octagon
(G) pentagon
(H) rectangle
(I) sector
(J) triangle \\
\midrule
Ground truth Answer & \textbf{(F) octagon} \\
\\
\midrule
\textbf{Method} & \textbf{Output} \\
\midrule
\midrule
Direct Prompting & 
\begin{tabular}[t]{@{}l@{}}
Final answer: (C) hexagon \textbf{(Wrong)}
\end{tabular} \\
\midrule
Zero-shot Prompting & 
\begin{tabular}[t]{@{}l@{}}
The given SVG path forms a shape with 8 sides, which is a hexagon. \\
Therefore, the answer is (C) hexagon. \textbf{(Wrong)}
\end{tabular} \\
\midrule
PoT & 
\begin{tabular}[t]{@{}l@{}}
def solution(): \\
\hspace*{1em}return "C) hexagon" \\
\\
solution() \\
\texttt{>>>} C) hexagon \textbf{(Wrong)}
\end{tabular} \\
\midrule
NL Planning & 
\begin{tabular}[t]{@{}l@{}}
Step 1: Identify the SVG Path Element. The given SVG path element is \\ \texttt{<}path d="M 38.00,62.00 L 48.00,60.00 L 51.00,49.00 L 54.00,60.00 L 65.00,\\62.00 L 54.00,64.00 L 51.00,74.00 L 48.00,64.00 L 38.00,62.00"\texttt{/>} \\
\\
Step 2: Analyze the Path Data. The path data commands used are 'M' \\for move to and 'L' for line to. The sequence of points mentioned are \\(38.00,62.00), (48.00,60.00), (51.00,49.00), (54.00,60.00), (65.00,62.00), \\(54.00,64.00), (51.00,74.00), (48.00,64.00), (38.00,62.00). \\
\\
Step 3: Determine the Shape. Based on the sequence of points and \\movements, the shape drawn appears to be a kite.\\
\\
Step 4: Match with Options. The shape drawn matches with option (D) kite. \\
\\
Step 5: Output the Correct Option. \\
Final answer: (D) kite \textbf{(Wrong)}
\end{tabular} \\
\midrule
Ours & 
\begin{tabular}[t]{@{}l@{}}
Extracted path commands: ['M 38.00,62.00', 'L 48.00,60.00', 'L 51.00,49.00', \\'L 54.00,60.00', 'L 65.00,62.00', 'L 54.00,64.00', 'L 51.00,74.00', 'L 48.00,64.00', \\'L 38.00,62.00'] \\
Executing command: M 38.00,62.00 - Action: Move to (38.00, 62.00) \\- New coordinates: (38.00, 62.00) \\
Executing command: L 48.00,60.00 - Action: Draw line to (48.00, 60.00) \\- New coordinates: (48.00, 60.00) \\
Executing command: L 51.00,49.00 - Action: Draw line to (51.00, 49.00) \\- New coordinates: (51.00, 49.00) \\
Executing command: L 54.00,60.00 - Action: Draw line to (54.00, 60.00) \\- New coordinates: (54.00, 60.00) \\
Executing command: L 65.00,62.00 - Action: Draw line to (65.00, 62.00) \\- New coordinates: (65.00, 62.00) \\
Executing command: L 54.00,64.00 - Action: Draw line to (54.00, 64.00) \\- New coordinates: (54.00, 64.00) \\
Executing command: L 51.00,74.00 - Action: Draw line to (51.00, 74.00) \\- New coordinates: (51.00, 74.00) \\
Executing command: L 48.00,64.00 - Action: Draw line to (48.00, 64.00) \\- New coordinates: (48.00, 64.00) \\
Executing command: L 38.00,62.00 - Action: Draw line to (38.00, 62.00) \\- New coordinates: (38.00, 62.00) \\
Identified shape: octagon \\
Final answer: (F) octagon \textbf{(Correct)}
\end{tabular} \\
\bottomrule
\end{tabularx}
\centering
\caption{A comparison of results for Geometric Shapes between the baseline methods and \ours.}
\label{tab:geometric_comparision}
\end{table}
\begin{table}[ht]
\small
\begin{tabularx}{\textwidth}{lX}
\toprule
Example Instance & If you follow these instructions, do you return to the starting point? Take 3 steps. Turn around. Take 5 steps. Turn right. Turn right. Take 1 step. Take 1 step. \\
\midrule
Ground truth Answer & \textbf{Yes} \\
\\
\midrule
\textbf{Method} & \textbf{Output} \\
\midrule
\midrule
Direct Prompting & 
\begin{tabular}[t]{@{}l@{}}
Final answer: no \textbf{(Wrong)}
\end{tabular} \\
\midrule
Zero-shot Prompting & 
\begin{tabular}[t]{@{}l@{}}
First, take 3 steps. Then, turn around and take 5 steps in the opposite \\direction. After that, turn right twice and take 1 step followed by \\another step. \\
Therefore, the answer is no. \textbf{(Wrong)}
\end{tabular} \\
\midrule
PoT & 
\begin{tabular}[t]{@{}l@{}}
def solution(): \\
\hspace*{1em}x = 0 \\
\hspace*{1em}y = 0 \\
\hspace*{1em}direction = 0  \# 0: North, 1: East, 2: South, 3: West \\
\\
\hspace*{1em}\# Take 3 steps \\
\hspace*{1em}for \_ in range(3): \\
\hspace*{2em}if direction == 0: \\
\hspace*{3em}y += 1 \\
\hspace*{2em}elif direction == 1: \\
\hspace*{3em}x += 1 \\
\hspace*{2em}elif direction == 2: \\
\hspace*{3em}y -= 1 \\
\hspace*{2em}else: \\
\hspace*{3em}x -= 1 \\
\\
\hspace*{1em}\# Turn around \\
\hspace*{1em}direction = (direction + 2) \% 4 \\
\\
\hspace*{1em}\# Take 5 steps \\
\hspace*{1em}... \\
\\
\hspace*{1em}return x == 0 and y == 0 \\
solution() \\
\texttt{>>>} True \textbf{(Wrong)}
\end{tabular} \\
\midrule
NL Planning & 
\begin{tabular}[t]{@{}l@{}}
Final answer: No \textbf{(Wrong)}
\end{tabular} \\
\midrule
Ours & 
\begin{tabular}[t]{@{}l@{}}
Instructions: ['Take 3 steps', 'Turn around', 'Take 5 steps',\\ 'Turn right', 'Turn right', 'Take 1 step', 'Take 1 step'] \\
Initial position: (0, 0), Orientation: N \\
Instruction: Take 3 steps \texttt{->} Position: (0, 3), Orientation: N \\
Instruction: Turn around \texttt{->} Position: (0, 3), Orientation: S \\
Instruction: Take 5 steps \texttt{->} Position: (0, -2), Orientation: S \\
Instruction: Turn right \texttt{->} Position: (0, -2), Orientation: W \\
Instruction: Turn right \texttt{->} Position: (0, -2), Orientation: N \\
Instruction: Take 1 step \texttt{->} Position: (0, -1), Orientation: N \\
Instruction: Take 1 step \texttt{->} Position: (0, 0), Orientation: N \\
Final Position: (0, 0) \\
Final answer: Yes \textbf{(Correct)}
\end{tabular} \\
\bottomrule
\end{tabularx}
\centering
\caption{A comparison of results for Navigate between the baseline methods and \ours.}
\label{tab:navigate_comparison}
\end{table}
\begin{table}[ht]
\tiny
\begin{tabularx}{\textwidth}{lX}
\toprule
Example Instance & On the floor, there is one mauve cat toy, two purple cat toys, three grey cat toys, two mauve notebooks, three grey notebooks, three burgundy cat toys, and one purple notebook. If I remove all the notebooks from the floor, how many grey objects remain on it? Options: (A) zero (B) one (C) two (D) three (E) four (F) five (G) six (H) seven (I) eight (J) nine (K) ten (L) eleven (M) twelve (N) thirteen (O) fourteen (P) fifteen (Q) sixteen \\
\midrule
Ground truth Answer & (D) three \\
\midrule
\textbf{Method} & \textbf{Output} \\
\midrule
\midrule
Direct Prompting & Final answer: (C) two \textbf{(Wrong)} \\
\midrule
Zero-shot Prompting & First, let's count the number of grey objects on the floor before removing the notebooks: - Three grey cat toys - Three grey notebooks Total grey objects before removing notebooks: 3 (grey cat toys) + 3 (grey notebooks) = 6 grey objects Now, let's remove the notebooks from the floor: - Two mauve notebooks - Three grey notebooks - One purple notebook Total notebooks removed: 2 (mauve notebooks) + 3 (grey notebooks) + 1 (purple notebook) = 6 notebooks Now, let's calculate how many grey objects remain on the floor after removing the notebooks: Total grey objects before removing notebooks: 6 grey objects Total notebooks removed: 6 notebooks Grey objects remaining on the floor: 6 (total grey objects) - 6 (notebooks removed) = 0 grey objects Therefore, the answer is (A) zero \textbf{(Wrong)} \\
\midrule
PoT & def solution(): \newline \hspace*{1em} \# Total number of grey objects on the floor \newline \hspace*{1em} total\_grey\_objects = 3 + 3 \newline \newline \hspace*{1em} \# Removing notebooks from the floor \newline \hspace*{1em} removed\_notebooks = 2 + 3 \newline \newline \hspace*{1em} \# Number of grey objects remaining on the floor \newline \hspace*{1em} remaining\_grey\_objects = total\_grey\_objects - removed\_notebooks \newline \newline \hspace*{1em} return remaining\_grey\_objects \newline \newline solution() \newline \texttt{>>>} 1 \textbf{(Wrong)} \\
\midrule
NL Planning & Step 1: - Mauve cat toy - Purple cat toys - Grey cat toys - Mauve notebooks - Grey notebooks - Burgundy cat toys - Purple notebook Step 2: - The question asks about the number of grey objects remaining on the floor after removing all the notebooks. Step 3: - There are three grey cat toys and three grey notebooks on the floor. Step 4: - After removing all the notebooks, there are three grey cat toys remaining. Step 5: - The answer to the question is (C) two. Final answer: (C) two \textbf{(Wrong)} \\
\midrule
Ours & Items on the surface: { 'mauve cat toy': 1, 'purple cat toy': 2, 'grey cat toy': 3, 'mauve notebook': 2, 'grey notebook': 3, 'burgundy cat toy': 3, 'purple notebook': 1} Spatial relationships between items: { 'mauve cat toy': ['floor'], 'purple cat toy': ['floor'], 'grey cat toy': ['floor'], 'mauve notebook': ['floor'], 'grey notebook': ['floor'], 'burgundy cat toy': ['floor'], 'purple notebook': ['floor']} The question is: how many grey objects remain on it? Reasoning for the answer: After removing all the notebooks, only the grey cat toys remain on the floor. Answer options: [ '(A) zero', '(B) one', '(C) two', '(D) three', '(E) four', '(F) five', '(G) six', '(H) seven', '(I) eight', '(J) nine', '(K) ten', '(L) eleven', '(M) twelve', '(N) thirteen', '(O) fourteen', '(P) fifteen', '(Q) sixteen'] Final answer: (D) three \textbf{(Correct)} \\
\bottomrule
\end{tabularx}
\caption{A comparison of results for Reasoning about Colored Objects Shapes between the baseline methods and ours.}
\label{tab:reasoning_comparison}
\end{table}

\begin{table}[ht]
\tiny
\begin{tabularx}{\textwidth}{lX}
\toprule
Example Instance & Today, Jason went to the movies. Between what times could they have gone? We know that: Jason woke up at 10am.
Linda saw Jason getting a coffee at the cafe from 10am to 3pm.
Jennifer saw Jason walking towards the Statue of Liberty from 6pm to 7pm.
Sean saw Jason buying a phone at the electronics store from 7pm to 10pm.
The movies was closed after 10pm.
Between what times could Jason have gone to the movies?
Options:
(A) 3pm to 6pm
(B) 10am to 3pm
(C) 7pm to 10pm
(D) 6pm to 7pm \\
\midrule
Ground truth Answer & \textbf{(A) 3pm to 6pm} \\
\\
\midrule
\textbf{Method} & \textbf{Output} \\
\midrule
\midrule
Direct Prompting & 
\begin{tabular}[t]{@{}l@{}}
Final answer: (C) 7pm to 10pm \textbf{(Wrong)}
\end{tabular} \\
\midrule
Zero-shot Prompting & 
\begin{tabular}[t]{@{}l@{}}
Jason could have gone to the movies between 7pm to 10pm, as the \\movies were closed after 10pm.\\
Therefore, the answer is (C) 7pm to 10pm \textbf{(Wrong)}
\end{tabular} \\
\midrule
PoT & 
\begin{tabular}[t]{@{}l@{}}
def solution(): \\
\hspace*{1em}possible\_times = [] \\
\\
\hspace*{1em}\# Jason could have gone to the movies between 3pm to 6pm \\
\hspace*{1em}if 3 <= 6: \\
\hspace*{2em}possible\_times.append("3pm to 6pm") \\
\\
\hspace*{1em}\# Jason could have gone to the movies between 10am to 3pm \\
\hspace*{1em}if 10 <= 3: \\
\hspace*{2em}possible\_times.append("10am to 3pm") \\
\\
\hspace*{1em}\# Jason could have gone to the movies between 7pm to 10pm \\
\hspace*{1em}if 7 <= 10: \\
\hspace*{2em}possible\_times.append("7pm to 10pm") \\
\\
\hspace*{1em}\# Jason could have gone to the movies between 6pm to 7pm \\
\hspace*{1em}if 6 <= 7: \\
\hspace*{2em}possible\_times.append("6pm to 7pm") \\
\\
\hspace*{1em}return possible\_times \\
\\
solution() \\
\texttt{>>>} ['3pm to 6pm', '7pm to 10pm', '6pm to 7pm'] \textbf{(Wrong)}
\end{tabular} \\
\midrule
NL Planning & 
\begin{tabular}[t]{@{}l@{}}
Step 1: \\
- Jason woke up at 10am.\\
- Linda saw Jason at the cafe from 10am to 3pm.\\
- Jennifer saw Jason walking towards the Statue of Liberty from 6pm to 7pm.\\
- Sean saw Jason at the electronics store from 7pm to 10pm.\\
- The movies was closed after 10pm.\\
Step 2:\\
- The specific question is about the time range for Jason to have gone to the movies.\\
Step 3:\\
- Timeline:\\
  - 10am: Jason woke up\\
  - 10am to 3pm: Jason at the cafe\\
  - 6pm to 7pm: Jason walking towards the Statue of Liberty\\
  - 7pm to 10pm: Jason at the electronics store\\
  - After 10pm: Movies closed\\
Step 4:\\
- Events placed on the timeline accordingly.\\
Step 5:\\
- Possible time range for Jason to have gone to the movies is after 10pm.\\
Step 6:\\
- Comparison results:\\
  - Option (A) 3pm to 6pm: Not possible\\
  - Option (B) 10am to 3pm: Not possible\\
  - Option (C) 7pm to 10pm: Possible\\
  - Option (D) 6pm to 7pm: Not possible\\
Step 7:\\
- Final answer: Jason could have gone to the movies between 7pm to 10pm. (Option C) \textbf{(Wrong)}
\end{tabular} \\
\midrule
Ours & 
\begin{tabular}[t]{@{}l@{}}
Events and their timings: [('Jason woke up', '10am'), ('Linda saw Jason getting a coffee at the cafe', '10am to 3pm'),\\ ('Jennifer saw Jason walking towards the Statue of Liberty', '6pm to 7pm'), \\('Sean saw Jason buying a phone at the electronics store', '7pm to 10pm')]\\
Conflicting time intervals: [('10am', '3pm'), ('6pm', '7pm'), ('7pm', '10pm')]\\
Available time intervals for visitation: [('3pm', '6pm')]\\
Final answer: (A) 3pm to 6pm \textbf{(Correct)}
\end{tabular} \\
\bottomrule
\end{tabularx}
\centering
\caption{A comparison of results for Temporal Sequences between the baseline methods and \ours.}
\label{tab:temporal_comparison}
\end{table}
\begin{table}[ht]
\tiny
\begin{tabularx}{\textwidth}{lX}
\toprule
Example Instance & Alice, Bob, Claire, Dave, and Eve are playing a game. At the start of the game, they are each holding a ball: Alice has a orange ball, Bob has a black ball, Claire has a pink ball, Dave has a white ball, and Eve has a red ball.
As the game progresses, pairs of players trade balls. First, Alice and Dave swap balls. Then, Bob and Claire swap balls. Then, Claire and Dave swap balls. Then, Dave and Bob swap balls. Finally, Alice and Eve swap balls. At the end of the game, Claire has the
Options:
(A) orange ball
(B) black ball
(C) pink ball
(D) white ball
(E) red ball \\
\midrule
Ground truth Answer & \textbf{(A) orange ball} \\
\\
\midrule
\textbf{Method} & \textbf{Output} \\
\midrule
\midrule
Direct Prompting & 
\begin{tabular}[t]{@{}l@{}}
Final answer: (D) white ball \textbf{(Wrong)}
\end{tabular} \\
\midrule
Zero-shot Prompting & 
\begin{tabular}[t]{@{}l@{}}
1. After Alice and Dave swap balls:\\
Alice has a white ball,\\
Bob has a black ball,\\
Claire has a pink ball,\\
Dave has an orange ball,\\
Eve has a red ball.\\
\\
2. After Bob and Claire swap balls:\\
...\\
\\
5. After Alice and Eve swap balls:\\
Alice has a red ball,\\
Bob has an orange ball,\\
Claire has a black ball,\\
Dave has a pink ball,\\
Eve has a white ball.\\
\\
Therefore, the answer is (B) black ball \textbf{(Wrong)}
\end{tabular} \\
\midrule
PoT & 
\begin{tabular}[t]{@{}l@{}}
def solution(): \\
\hspace*{1em}balls = \{"Alice": "orange", "Bob": "black", "Claire": "pink", \\"Dave": "white", "Eve": "red"\} \\
\\
\hspace*{1em}\# Define the swaps \\
\hspace*{1em}swaps = [("Alice", "Dave"), ("Bob", "Claire"), ("Claire", "Dave"), \\("Dave", "Bob"), ("Alice", "Eve")] \\
\\
\hspace*{1em}\# Perform the swaps \\
\hspace*{1em}for swap in swaps: \\
\hspace*{2em}temp = balls[swap[0]] \\
\hspace*{2em}balls[swap[0]] = balls[swap[1]] \\
\hspace*{2em}balls[swap[1]] = temp \\
\\
\hspace*{1em}\# Return the ball color Claire has at the end \\
\hspace*{1em}return [key for key, value in balls.items() if value == "pink"][0] \\
\\
solution() \\
\texttt{>>>} Dave \textbf{(Wrong)}
\end{tabular} \\
\midrule
NL Planning & 
\begin{tabular}[t]{@{}l@{}}
Step 1: \\
Alice - orange ball\\
Bob - black ball\\
Claire - pink ball\\
Dave - white ball\\
Eve - red ball\\
\\
Step 2:\\
1. Alice and Dave swap balls\\
2. Bob and Claire swap balls\\
3. Claire and Dave swap balls\\
4. Dave and Bob swap balls\\
5. Alice and Eve swap balls\\
\\
Step 3:\\
Alice - white ball\\
Bob - pink ball\\
Claire - black ball\\
Dave - black ball\\
Eve - orange ball\\
\\
Step 4:\\
Final book assignment for Claire: black ball\\
\\
Step 5:\\
Options: ...\\

Step 6:
Final answer: (B) black ball \textbf{(Wrong)}
\end{tabular} \\
\midrule
Ours & 
\begin{tabular}[t]{@{}l@{}}
Initial assignments: {'Alice': 'orange ball', 'Bob': 'black ball', 'Claire': 'pink ball', 'Dave': 'white ball', 'Eve': 'red ball'}\\
Swap pairs: [('Alice', 'Dave'), ('Bob', 'Claire'), ('Claire', 'Dave'), ('Dave', 'Bob'), ('Alice', 'Eve')]\\
After swapping Alice and Dave: {'Alice': 'white ball', 'Bob': 'black ball', 'Claire': 'pink ball', 'Dave': 'orange ball', 'Eve': 'red ball'}\\
...\\
After swapping Alice and Eve: {'Alice': 'red ball', 'Bob': 'black ball', 'Claire': 'orange ball', 'Dave': 'pink ball', 'Eve': 'white ball'}\\
The final position or gift of Claire is: orange ball\\
Options: [('(A) orange ball', '(B) black ball', '(C) pink ball', '(D) white ball', '(E) red ball']\\
Final answer:(A) \textbf{(Correct)}
\end{tabular} \\
\bottomrule
\end{tabularx}
\centering
\caption{A comparison of results for Tracking Shuffled Objectives between the baseline methods and \ours.}
\label{tab:tracking_comparison}
\end{table}
\begin{table}[ht]
\tiny
\begin{tabularx}{\textwidth}{lX}
\toprule
Example Instance & Vina tells the truth. Helene says Vina lies. Kandi says Helene tells the truth. \\&Jamey says Kandi lies. Ka says Jamey lies. Does Ka tell the truth?\\
\midrule
Ground truth Answer & \textbf{No} \\
\\
\midrule
\textbf{Method} & \textbf{Output} \\
\midrule
\midrule
Direct Prompting & 
\begin{tabular}[t]{@{}l@{}}
Final answer: Yes \textbf{(Wrong)}
\end{tabular} \\
\midrule
Zero-shot Prompting & 
\begin{tabular}[t]{@{}l@{}}
First, let's analyze the statements:\\
1. Vina tells the truth.\\
2. Helene says Vina lies.\\
3. Kandi says Helene tells the truth.\\
4. Jamey says Kandi lies.\\
5. Ka says Jamey lies.\\
\\
From statement 2, we can infer that Helene lies, which means Vina \\actually tells the truth.\\
From statement 3, we can infer that Kandi tells the truth.\\
From statement 4, we can infer that Jamey lies.\\
From statement 5, we can infer that Ka tells the truth.\\
\\
Therefore, Ka tells the truth.\\
Therefore, the answer is Yes \textbf{(Wrong)}
\end{tabular} \\
\midrule
PoT & 
\begin{tabular}[t]{@{}l@{}}
def solution(): \\
\hspace*{1em}vina = True \\
\hspace*{1em}helene = not vina \\
\hspace*{1em}kandi = vina \\
\hspace*{1em}jamey = not kandi \\
\hspace*{1em}ka = not jamey \\
\\
\hspace*{1em}return ka \\
\\
solution() \\
\texttt{>>>} True \textbf{(Wrong)}
\end{tabular} \\
\midrule
NL Planning & 
\begin{tabular}[t]{@{}l@{}}
Step 1: \\
- Vina tells the truth. \\
- Helene says Vina lies. \\
- Kandi says Helene tells the truth. \\
- Jamey says Kandi lies. \\
- Ka says Jamey lies. \\

Step 2: \\
\begin{tabular}{|l|l|}
\hline
Person & Truth-telling status \\
\hline
Vina   & True \\
Helene & False \\
Kandi  & True \\
Jamey  & False \\
Ka     & False \\
\hline
\end{tabular} \\

Step 3: \\
- Based on Helene's statement, Vina is lying. Since Vina tells the truth, \\Helene is lying. \\
- Based on Kandi's statement, Helene tells the truth. Since Helene lies, \\Kandi is lying. \\
- Based on Jamey's statement, Kandi lies. Since Kandi tells the truth, \\Jamey is lying. \\
- Based on Ka's statement, Jamey lies. Since Jamey lies, Ka is telling \\the truth. \\

Step 4: \\
\begin{tabular}{|l|l|}
\hline
Person & Truth-telling status \\
\hline
Vina   & True \\
Helene & False \\
Kandi  & True \\
Jamey  & False \\
Ka     & True \\
\hline
\end{tabular} \\

Step 5: \\
- Ka tells the truth. \\

Final answer: Yes \textbf{(Wrong)}
\end{tabular} \\
\midrule
Ours & 
\begin{tabular}[t]{@{}l@{}}
Statements: ['Vina tells the truth', 'Helene says Vina lies', \\'Kandi says Helene tells the truth', \\'Jamey says Kandi lies', 'Ka says Jamey lies']\\
Question:  Does Ka tell the truth\\
Vina says the truth. Vina tells the truth: True\\
Helene says Vina lies. Helene tells the truth: False\\
Kandi says Helene tells the truth. Kandi tells the truth: False\\
Jamey says Kandi lies. Jamey tells the truth: True\\
Ka says Jamey lies. Ka tells the truth: False\\
Final answer: No \textbf{(Correct)}
\end{tabular} \\
\bottomrule
\end{tabularx}
\centering
\caption{A comparison of results for Web of lies between the baseline methods and \ours.}
\label{tab:web_comparison}
\end{table}
\end{document}